\documentclass[10pt,journal,compsoc]{IEEEtran}

\usepackage{soul}
\usepackage{url}
\usepackage{booktabs}
\usepackage{amsmath}
\usepackage{amsthm} 
\usepackage{amsfonts}  
\usepackage{amssymb}
\usepackage{bm}
\usepackage{multicol}
\usepackage{multirow}
\usepackage{graphicx}
\usepackage{booktabs}
\usepackage{colortbl}
\usepackage{stfloats}
\usepackage{comment}
\usepackage{makecell}
\usepackage[noend]{algpseudocode}
\usepackage{algorithmicx,algorithm}
\usepackage{pifont}
\usepackage{xcolor}
\usepackage{color}
\usepackage{microtype}
\usepackage{cleveref}
\usepackage[]{footmisc}
\usepackage{paralist}
\usepackage{caption}
\usepackage{arydshln}
\usepackage{tabularx}

\usepackage{wrapfig}
\usepackage{tikz}
\usepackage{ragged2e}

\definecolor{finsta-color}{RGB}{203,0,180}
\definecolor{blue}{RGB}{218,232,252}
\definecolor{mblue}{RGB}{23,0,255}
\definecolor{mred}{RGB}{214,59,44}

\definecolor{nmred}{RGB}{255,229,229}
\definecolor{nmdeepred}{RGB}{255,183,183}
\definecolor{nmgreen}{RGB}{196,237,227}
\definecolor{nmdeepgreen}{RGB}{126,227,203}
\definecolor{nmyellow}{RGB}{255,242,216}
\definecolor{nmdeepyellow}{RGB}{250,224,167}

\ifCLASSOPTIONcompsoc
  \usepackage[nocompress]{cite}
\else
  \usepackage{cite}
\fi

\newcommand{\specialcell}[2][c]{\begin{tabular}[#1]{@{}c@{}}#2\end{tabular}}

\newcommand{\halfcheckmark}{\checkmark\hspace{-2.5mm}\raisebox{0.8mm}{\large\rotatebox{-40}{-}}\hspace{1mm}}

\begin{document}

\title{Enhancing Video-Language Representations\\ with Structural Spatio-Temporal Alignment}

\author{Hao Fei,~\IEEEmembership{Member,~IEEE}, Shengqiong Wu, Meishan Zhang, \protect\\ 
Min Zhang, Tat-Seng Chua and~Shuicheng Yan,~\IEEEmembership{Fellow,~IEEE}
\IEEEcompsocitemizethanks{\IEEEcompsocthanksitem Hao Fei, Shengqiong Wu and Tat-Seng Chua are with School of Computing, National University of Singapore, Singapore.\protect\\
E-mail: \{haofei37, dcscts\}@nus.edu.sg, swu@u.nus.edu
\IEEEcompsocthanksitem Meishan Zhang and Min Zhang are with Harbin Institute of Technology (Shenzhen), China. (Corresponding author: Meishan Zhang.) \protect\\
E-mail: \{zhangmeishan, zhangmin2021\}@hit.edu.cn
\IEEEcompsocthanksitem Shuicheng Yan is with Skywork AI, Kunlun 2050 Research, Singapore. \protect\\
E-mail: shuicheng.yan@kunlun-inc.com}
}

\IEEEtitleabstractindextext{%
\begin{abstract}\justifying
While pre-training large-scale video-language models (VLMs) has shown remarkable potential for various downstream video-language tasks, existing VLMs can still suffer from certain commonly seen limitations, e.g., \emph{coarse-grained cross-modal aligning}, 
\emph{under-modeling of temporal dynamics}, \emph{detached video-language view}.
In this work, we target enhancing VLMs with a \textbf{fine-grained structural spatio-temporal alignment learning} method (namely Finsta).
First of all, we represent the input texts and videos with fine-grained scene graph (SG) structures, both of which are further unified into a holistic SG (HSG) for bridging two modalities.
Then, an SG-based framework is built, where the textual SG (TSG) is encoded with a graph Transformer, while the video dynamic SG (DSG) and the HSG are modeled with a novel recurrent graph Transformer for spatial and temporal feature propagation.
A spatial-temporal Gaussian differential graph Transformer is further devised to strengthen the sense of the changes in objects across spatial and temporal dimensions.
Next, based on the fine-grained structural features of TSG and DSG, we perform object-centered spatial alignment and predicate-centered temporal alignment respectively, enhancing the video-language grounding in both the spatiality and temporality.
We design our method as a plug\&play system, which can be integrated into existing well-trained VLMs for further representation augmentation, without training from scratch or relying on SG annotations in downstream applications.
On 6 representative VL modeling tasks over 12 datasets in both standard and long-form video scenarios, Finsta consistently improves the existing 13 strong-performing VLMs persistently, and refreshes the current state-of-the-art end task performance significantly in both the fine-tuning and zero-shot settings.
\end{abstract}

}

\maketitle

\IEEEdisplaynontitleabstractindextext

\IEEEpeerreviewmaketitle

\IEEEraisesectionheading{\section{Introduction}\label{Introduction}}
\IEEEPARstart{R}ecently, large language model (LLM) pre-training over various data modalities (e.g., texts, images and videos) has shown amazing potential in ushering human-level intelligence, such as GPT4 \cite{bubeck2023gpt4}, PaLM-E~\cite{driess2023palme}, BLIP-2~\cite{li2023blip2}, Flamingo~\cite{AlayracDLMBHLMM22}, LLaVA~\cite{abs-2304-08485}.
Among them, the topic of video-language model (VLM) pre-training has received an increasing number of research attention \cite{abs-2002-06353,abs-2207-07885,JuHZZX22,maaz2023video,lin2023video}.
Compared with vision-text modeling which focuses mainly on the individual visual semantic understanding, video understanding goes beyond static images, requiring both the comprehension of spatial semantics and temporal dynamics, due to the nature of a sequence of frames over time.
Extensive efforts have been made for learning effective VLMs, and facilitated a broad range of downstream video-language (VL) tasks \cite{Wang0WLZZX0JY22}.

Despite the promising progress, existing VLMs could be still subject to certain common yet crucial issues that are intrinsically raised by the nature of video-text modality heterogeneity.
Consequently, the performance of downstream VL tasks might still be hampered from achieving optimal.

\begin{figure*}[!t]
\centering
\includegraphics[width=0.98\textwidth]{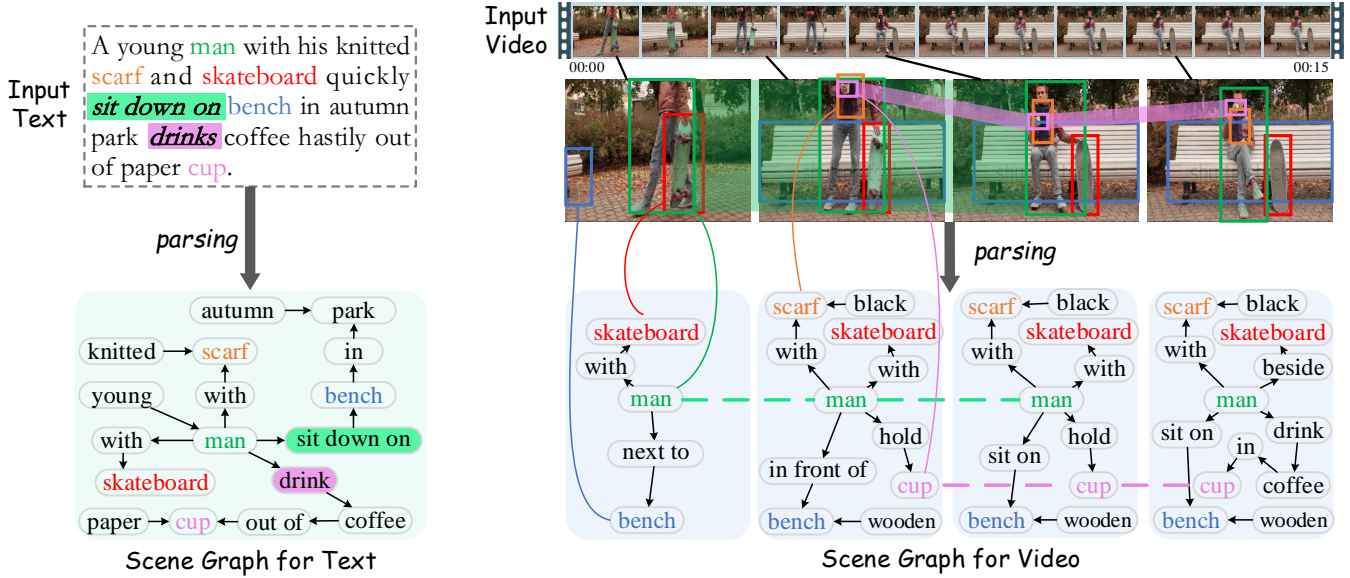}
\vspace{-2mm}
\caption{
Representing the text and video with the corresponding scene graphs (SGs) enables more fine-grained control of video-language correspondence learning (same colors denote same concepts), as the SG representations depict the intrinsic modal-agnostic semantic structures of texts or videos.
Best viewed in color.
}
\vspace{-4mm}
\label{intro}
\end{figure*}

\begin{compactitem}
\item \textbf{First, coarse-grained cross-modal aligning}.
Current works extensively perform alignment either between the overall video and text representations \cite{LuoJZCLDL22,ye2023hitea}, or extracting frame patches \cite{xu-etal-2021-videoclip,LeiLZGBB021}.
Nevertheless, these two modalities are unequal in carrying information, e.g., texts are limited and succinct while videos encompass dense surplus contents, which inevitably leads to low-effective alignment when using a coarse-grained manner.
For example, in video captioning, this can lead to generating brief captions without sufficient detail.
As highlighted in Figure \ref{intro}, the correspondences between video and text can be actually fine-grained, i.e., the co-referred object of interest.

\item \textbf{Second, under-modeling of temporal dynamics}.
The language uses abstract words (e.g., predicates and adverbial modifiers) to express complicated actions; while videos describe the events with dynamic changing of specific scenes in consecutive frames.
For example in Figure \ref{intro}, the text action `\emph{sit down on}' is depicted by the tracking process over the \emph{man} object in the video.
This thus suggests a delicate motion-level alignment to model the temporal dynamics correspondences of video-text data.
Unfortunately, existing research mostly takes the straightforward manner of video temporality modeling, i.e., with temporal attention or pooling over the overall frames \cite{SunXS00F22,chen-etal-2022-litevl}.
Consequently, e.g., for the video temporal localization task where modeling temporal dynamics is required, VLMs largely fail to accurately match textual content with the corresponding video temporal content.

\item \textbf{Last, detached video-language view}.
While two modalities come with shared features, there can be also rich differentiated information.
Intuitively, texts provide abstract expressions (e.g., emotions\&feelings) and videos are visual-sensible signals (e.g., colors\&appearances).
Such distinctions can serve a complementary role to each other for a holistic multimodal semantic understanding.
Yet current work pays the focus exclusively on VL alignment, while mostly treating the unaligned part of features as noises, and aggressively discards them without modeling \cite{abs-2002-06353,XueHZS00FG22}.
In reasoning-intensive scenarios, e.g., video question answering, VLMs cannot fully leverage both modality-sharing and modality-complementary information for in-depth reasoning.

\end{compactitem}

We argue that the fine-grained structured representations of video and text are indispensable for comprehensive VL understanding, as we human beings always grasp the underlying semantic structure before the next reasoning over VL.
In this work, we consider representing the input video and text with the scene graph (SG) representations \cite{JohnsonKSLSBL15}.
As shown in Figure \ref{intro}, depicting the intrinsic semantic relations of contents in texts or videos with structured modal-agnostic representations,
SGs enable fine-grained control of VL learning.
Based on the SG representations, we further make certain customizations before using it for our purpose.
First, we slightly retrofit the existing textual SG (TSG) definition \cite{SchusterKCFM15} by adding a predicate attribute node to further support the \emph{adverbial modifiers} of actions (e.g., `\emph{quickly}', `\emph{hastily}'), such that the TSG expression for motions can be augmented.
Second, for the dynamic SG (DSG) of video \cite{JiK0N20}, we connect the SG sequence of DSG as a whole by creating a type of \emph{temporal coreference edges} across different DSG frames.
Finally, we unify the TSG and DSG with a type of \emph{cross-modal coreference edges}, resulting in an overall holistic SG (HSG) of both video and language.
Via HSG (cf. Figure \ref{HSG}), we are able to secure a comprehensive view of multimodal semantics.

\vspace{-1mm}
Built upon the TSG, DSG and HSG, we then propose a \underline{\bf f}ine-gra\underline{\bf in}ed structural \underline{\bf s}patio-\underline{\bf t}emporal \underline{\bf a}lignment learning (namely \textbf{Finsta}) framework.
Our framework has a dual-stream-sum architecture, as shown in Figure \ref{framework1}.
We first adopt a graph Transformer (GTrm) \cite{abs-2012-09699} model for the highly-parallel graph encoding of TSG.
Based on GTrm we then devise a novel recurrent graph Transformer (R-GTrm) for spatio-temporal propagations of DSG and HSG.
We further propose a spatial-temporal Gaussian differential graph Transformer (STGD-GTrm) for strengthening the perception of the changes in objects across spatial and temporal dimensions, which learns to differentiate between moving nodes and stationary nodes. 
Next, based on the structural features we perform fine-grained VL alignment learning with respect to both spatiality and temporality.
Specifically, a \emph{high-order object-centered spatial contrastive} (OSC) learning and a \emph{high-order predicate-centered temporal contrastive} (PTC) learning are introduced for the cross-modal alignment learning between the TSG and DSG.
Finally, we present a representation transfer mechanism, in which we inject the well-aligned VL feature representations into a host VLM, which can be any existing VLMs.
With such plug-and-play design, we can enhance the VL representations of any well-trained VLMs by post-training with our Finsta module without the expensive pre-training from scratch; 
meanwhile avoiding the noise introduction of SG parsing when applying to downstream applications.

Extensive experiments are conducted on 6 representative video-language modeling tasks over 12 datasets in both standard and long-form video scenarios, such as Video Action Recognition \cite{feichtenhofer2016convolutional}, Video Captioning \cite{KrishnaHRFN17}, Video-Text Retrieval \cite{ChenZJW20}, Video Question Answering \cite{XuZX0Z0Z17}, Video-Paragraph Retrieval \cite{ren2023testa} and Long-form Video Question Answering \cite{SunXS00F22}.
Results show that our Finsta framework consistently improves the existing 10 strong-performing VLMs and 3 recent LVLMs persistently, and helps push new state-of-the-art VL end tasks significantly in both the fine-tuning and zero-shot settings.
Via further analyses, we verify that the proposed method effectively addresses the aforementioned bottlenecks of the VL learnings, including coarse-grained cross-modal aligning, under-modeling of temporal dynamics and insufficient VL collaboration. 
We also show empirical analyses to quantify the contributions of each module in Finsta, and explore the effects of a range of potential factors. 
Finally, we discuss the system's efficiency, and offer a collection of case studies to provide a direct insight into how Finsta's advancements unfold.

In summary, this work contributes in four key aspects:
\begin{compactitem}
    \item To our knowledge, we are the first to comprehensively enhance the video-language representation learning with structured fine-grained spatio-temporal alignment learning based on SG representations.
    
    \item Based on the GTrm, we devise a novel R-GTrm model for the spatial-temporal feature encoding of video.
    We further propose an STGD-GTrm to strengthen the perception of the changes in objects across spatial and temporal dimensions, differentiating between nodes in moving or stationary.

    \item We propose novel high-order object-centered spatial contrastive and high-order predicate-centered temporal contrastive learning strategies to realize the fine-grained spatio-temporal cross-modal alignment.

    \item Our method empirically boosts the current state-of-the-art VLMs on a broad range of downstream VL understanding tasks. 
    Besides, our framework is designed as a plug-and-play module, which can be easily applied to many existing VLMs broadly.
    
\end{compactitem}

\vspace{-3mm}
\section{Related Works}\label{Related Work}

\subsection{Video Language Modeling And Learning}

\vspace{-1mm}
Recently, large-scale language models have demonstrated the stunning ability of human-level semantics understanding \cite{driess2023palme}.
Such triumph of the text-based LLMs has soon expanded to multimodal models, such as GPT4 \cite{bubeck2023gpt4}, BLIP-2~\cite{li2023blip2} and Flamingo~\cite{AlayracDLMBHLMM22}, by further associating different multimodalities together, e.g., text, vision and video.
Among them, VLMs have received increasing research attention \cite{abs-2002-06353,abs-2207-07885,JuHZZX22}.
Building on the top of LLMs, video LLMs (i.e., Large VLMs, a.k.a. LVLMs) also have witnessed the rapid development, such as Video-LLaMA \cite{abs-2306-02858}, Video-ChatGPT \cite{maaz2023video}, Video-LLaVA \cite{lin2023video}.
VLM pre-training aims to learn a strong joint representation between two modalities by training models on large-scale video-text pairs.
VLM greatly facilitates a wide range of downstream VL understanding tasks, such as Video Captioning \cite{XuMYR16,SeoNAS22}, Video-Text Retrieval \cite{ChenZJW20} and Video Question Answering \cite{MaharajBRCP17,JangSYKK17}.
Aligned with the success of the prior pre-training of image-language models (ILMs)
\cite{HuangX0L21,RadfordKHRGASAM21,LeiLZGBB021}, some VLMs directly extend the cross-modal features of existing ILMs as the VLMs.
This is intuitive as both ILMs and VLMs can share much similar multimodal feature representations.
By taking existing ILMs as initiation, one can build a VLM with much lower costs, i.e., with fewer video-text pairs without training from scratch \cite{SunMV0S19}.

Based on such intuition, most of the existing VLMs take a similar idea of vision-text alignment learning as in ILM, where the major focus has been paid to the vision-language alignment \cite{HuangX0L21,RadfordKHRGASAM21,LuoJZCLDL22}.
However, videos can be essentially different from visions.
Compared with vision-text modeling which focuses mainly on the individual visual semantic understanding, video understanding goes beyond static images, requiring both the comprehension of spatial semantics and temporal dynamics, due to the nature of a sequence of frames over time.
Unfortunately, several key characteristics of the video-language modality heterogeneity have not been well considered in existing VLMs.
The first one is coarse-grained video-text alignment.
Existing VLMs mostly perform alignment either between the overall video and text representations \cite{LuoJZCLDL22,abs-2212-14546}, or extracting the regional frame patches \cite{xu-etal-2021-videoclip,LeiLZGBB021}.
Yet, texts and videos are unequal in carrying information, e.g., texts are limited and succinct while videos encompass dense surplus contents, which inevitably leads to low-effective alignment when using a coarse-grained manner.
Besides, the temporal dynamics is the pivotal crux of the video modality understanding, which however is not carefully modeled in existing VLMs.
As we stressed earlier, language has a natural gap with video.
Language often uses abstract words (e.g., predicates and adverbial modifiers) to express complicated actions; while videos describe the events with dynamic changing of specific scenes in consecutive frames.
Thus, a delicate motion-level alignment for the temporal dynamics correspondence modeling of video-text data is required.
In a word, a fine-grained consideration should be paid properly for both the spatial and temporal modeling of VLM.
To our knowledge, we are the first to model the fine-grained video-language spatio-temporal alignment by making use of SG representations.

\begin{figure*}[!t]
\centering
\includegraphics[width=0.95\textwidth]{illus/HSG7.pdf}
\caption{
We represent the input text and video with textual scene graph (TSG) and dynamic scene graph (DSG), respectively.
We unify the TSG and DSG into a holistic SG (HSG) by adding the cross-modal coreference edges.
}
\vspace{-3mm}
\label{HSG}
\end{figure*}

\vspace{-3mm}
\subsection{Scene Graph Representations}

\vspace{-1mm}
This study also closely relates to the application of scene graph representations \cite{JohnsonKSLSBL15}.
In SGs, the object and attribute nodes are connected in certain relations, where such graph structures intrinsically describe the underlying semantic meanings of input data.
With such characteristics, SGs have been extensively adopted as auxiliary features for downstream applications, e.g., Image Retrieval \cite{JohnsonKSLSBL15}, and Vision Captioning \cite{YangTZC19}.
In this work, we take advantage of SG representations, and perform cross-modal spatial-temporal alignment upon the SG structures.
We consider the textual SG \cite{SchusterKCFM15} to represent the language, and the dynamic SG \cite{JiK0N20} for the videos, both of which depict the modality-agnostic semantics.
We further construct a type of holistic SG over the TSG and DSG to maintain an overall unified VL representation.
To our knowledge, this is also the first attempt in the literature to connect the SG representations of both video and language.

In this paper, we also investigate the encoding approach of the fine-grained SG representations.
SG has naturally been depicted as a graph structure; thus the graph neural models can be used for the SG encoding, such as GCN \cite{marcheggiani-titov-2017-encoding}, GAT \cite{VelickovicCCRLB18}.
However, we consider the Transformer architecture \cite{VaswaniSPUJGKP17} for the graph modeling, as the self-attention calculation of the Transformer allows highly parallel computations.
And also the Transformer architecture has a more easy adaptation with the existing VLM backbone, such as ViT \cite{DosovitskiyB0WZ21}, VIOLET \cite{abs-2111-12681}, BEiT \cite{Bao0PW22} and more \cite{HuangX0L21,li2023blip2}.
We thus adopt the existing graph Transformer (GTrm) \cite{abs-2012-09699} to model the TSG structure.
On the other hand, the DSG has the temporal sequential characteristic.
Instead of using the RGNN \cite{VelickovicCCRLB18} for the DSG encoding, we newly devise a recurrent graph Transformer (R-GTrm).
We draw the main inspiration from the recurrent networks \cite{schuster1997bidirectional}.
Built upon the GTrm propagation, the video dynamics is additionally modeled in R-GTrm through the temporal coreference edges of nodes.

\vspace{-3mm}
\section{Scene Graph Construction}
\label{Construction of Scene Graphs}

\vspace{-1mm}

As aforementioned, we represent the input video and text with the DSG and TSG representations.
Following we describe the construction of the DSG and TSG of video and text, and also the HSG of a compound of both two.

\vspace{-2mm}
\subsection{Dynamic Scene Graph (DSG)}\label{constrction-DSG}
DSG describes a video into a temporally consecutive SGs.
Typically, each single visual SG comprises three types of nodes, including \emph{object}, \emph{attribute}, and \emph{relation} nodes.
As illustrated in Figure \ref{HSG}, the visual object nodes are connected in certain relations, and also objects are attached to their attributes.
The raw DSG maintains a visual SG for each video frame, while video frames are always redundant in content, and cause huge computation costs.
Thus, we first perform keyframe extraction for the video, such that the dense redundant video frames can be effectively compacted.
A clustering-based method \cite{song2016click} is used to extract the significant keyframes that faithfully keep the salient event contents in a proper sampling rate.
We record the raw time stamps $\tau_i$ of the resulting frames in the raw video, the key temporal information.
Then, these frames are fed into a parser for producing each static visual SG of each keyframe \cite{LiYX22}.

We follow the most common practice, and employ the FasterRCNN \cite{ren2015faster} as the object detector to obtain all the object nodes, where for each node $v^D_i$, we use 1) the object's neural representation $f_i$, 2) the object category label $c_i$ in the Visual Genome (VG) dataset, 3) the bounding box of the object $b_{i}$ (the 2D coordinate in the image, i.e., $(x^l,x^r,y^u,y^d)$).
And then we use MOTIFS as a relation classifier to obtain the relational edges as well as the relation labels.
We then use an attribute classifier to obtain attribute nodes.
All nodes (i.e., $v^D_i$ and $v^D_j$) are connected with edges $e^D_{i,j}$.

Since each single SG of DSG is separate in the graph sequence, we consider connecting them as a whole.
We create a type of \textbf{temporal coreference edges}, $e^D_{t-1\to t}$, for the objects across different SG frames, i.e., essentially a process of object tracking.
We realize this by measuring the intersection over union (IoU) of the bounding boxes ($b^{D,t}_i$ and $b^{D,t^{'}}_j$) of two objects with the same object label ($c^{D,t}_i$=$c^{D,t^{'}}_j$):
\begin{gather}
a^D(v^{D,t}_i,v^{D,t^{'}}_j) = \text{IoU}(b^{D,t}_i ,  b^{D,t^{'}}_j  \, | \, c^{D,t}_i = c^{D,t^{'}}_j ) \,.
\end{gather}
Essentially, the same objects across time always come with consecutive spatial movements in consistent labels.
We make comparisons with the object pairs between two consecutive SGs, and with the IoU value $a^D(v^{D,t}_i,v^{D,t^{'}}_j) > \gamma^D$ to be considered as the co-referred nodes, and we create a temporal coreference edge between these objects.
By ensembling the SGs of all the video keyframes via the temporal coreference edges, we obtain the resulting DSG ($\bm{\mathcal{G}}^D$) for the overall video,
as illustrated in Figure \ref{HSG}.
Formally,
\begin{gather}
\bm{\mathcal{G}}^D = \{\mathcal{G}^D_1,\cdots,\mathcal{G}^D_t \, | \, \mathcal{E}^D_{1\to 2},\cdots,E^D_{t-1\to t}\} \,, 
\end{gather}
where each SG frame $\mathcal{G}^{D}_t = (\mathcal{V}^{D}_t;\mathcal{E}^{D}_t)$, where:
\begin{gather}
\mathcal{V}^{D}_t = \{v^{D,t}_{i}\} = \{(f_i, c_i, b_{i}, \tau)^{D,t}\}  \,, \,\,\,
\mathcal{E}^{D}_t = \{e^{D,t}_{i,j}\}  \,,
\end{gather}
and each temporal coreference edge:
\begin{gather}
\mathcal{E}^{D}_{t-1\to t} = \{e^{D}_{1\to 2},\cdots,e^{D}_{t-1\to t}\}  \,.
\end{gather}

\subsection{Textual Scene Graph (TSG)}

The key difference between TSG and DSG lies in that TSG only comes with one single graph frame.
Similar to the visual SGs, TSGs also include three types of nodes, including \emph{object}, \emph{attribute}, and \emph{relation} nodes.
The objects are the textual entities within a scene, and each object has affiliate attributes connecting to it to describe the properties.
Note that the object nodes in visual SGs are images, while the object nodes in LSG are textual tokens, which are also the category labels of those objects.
So we only maintain the token/label $c^D_i$ as the node $v^D_i$.
And different types of nodes (i.e., $v^T_i$ and $v^T_j$) are connected with edges $e^T_{i,j}$.
Here relations could be either the persistent correlations (e.g., `\emph{in}', `\emph{with}' and `\emph{next to}') or some dynamic predicate words (e.g., `\emph{drink}', `\emph{sit down on}' and `\emph{hold}').
However, the raw TSG definition fails to support the \emph{adverbial modifiers} of the dynamic predicates.
For example, in the sentence of Figure \ref{intro}, the SG does not include the action modifiers `\emph{quickly}' for `\emph{drink}' and `\emph{hastily}' for `\emph{sit down on}'.
We note that this leads to important information loss, since in VL scenario, video can naturally depict such action states via its temporal characteristic.
Thus we introduce a type of dynamic attribute node of predicates, i.e., \emph{adverbial modifiers}.
We illustrate the retrofitted TSG in Figure \ref{HSG}.

In practice, we can obtain the TSG of texts via an off-the-shelf TSG parser \cite{SchusterKCFM15}.
We first convert the sentences into dependency trees with a dependency parser \cite{00010BT0GZ18}, which is then transformed into a graph based on the rules defined at \cite{SchusterKCFM15}.
To make TSG support the \textbf{adverbial modifiers} of the dynamic predicates, we retrofit the existing TSG parser, and keep the adverbial words or phases within the dependency tree, such that the TSG adds a type of attribute node of predicates.
Formally, we denote the resulting TSG as $\mathcal{G}^T$:
\begin{gather}
\mathcal{G}^T = (\mathcal{V}^{T};\mathcal{E}^{T}) \,, 
\end{gather}
where
\begin{gather}
\mathcal{V}^{T} = \{v^{T}_{i}\} = \{(c_i)^{T}\}  \,, \,\,\,\,
\mathcal{E}^{T} = \{e^{T}_{i,j}\}  \,.
\end{gather}

\begin{figure*}[!t]
\centering
\includegraphics[width=0.95\textwidth]{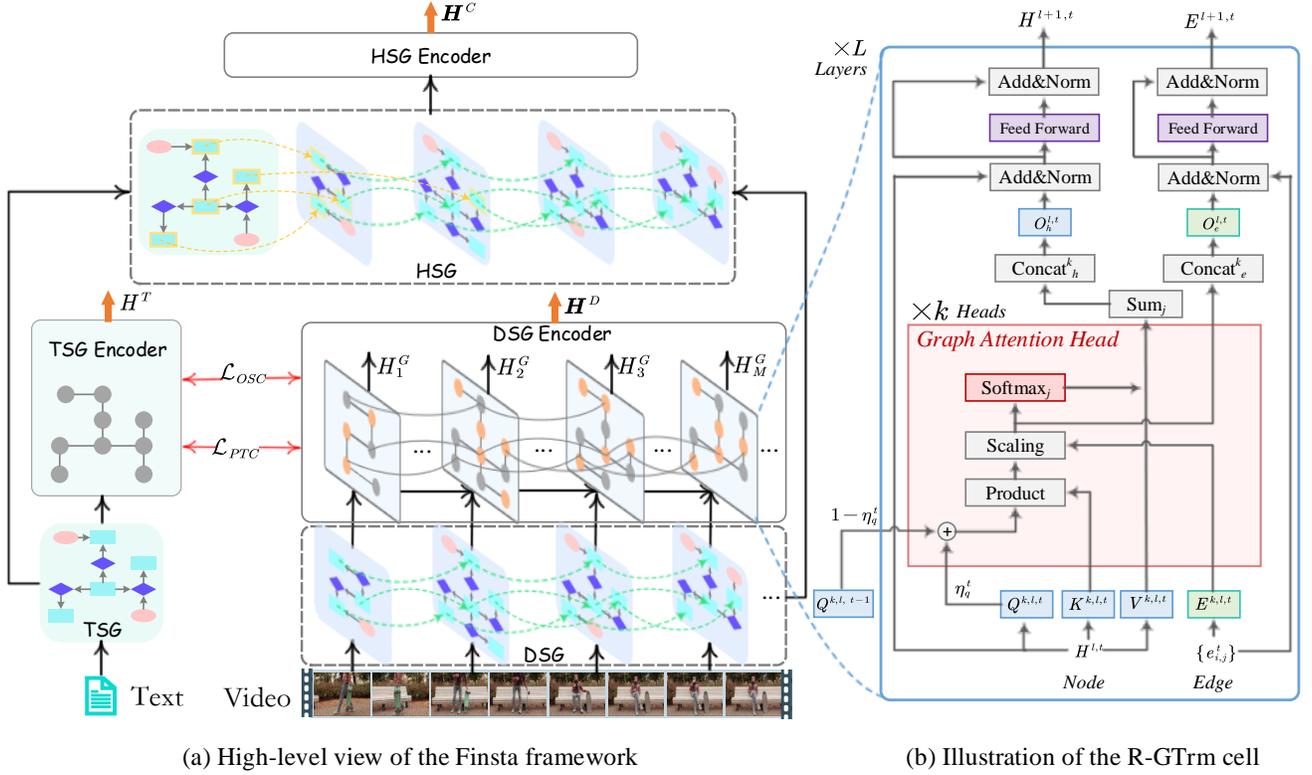}
\caption{
(a) The high-level view of our fine-grained structural spatio-temporal alignment learning (Finsta) framework based on the dual-stream-sum architecture.
And (b) the detailed dataflow of the recurrent graph Transformer (R-GTrm).
}
\vspace{-3mm}
\label{framework1}
\end{figure*}

\subsection{Holistic Scene Graph (HSG)}

Given a text-video pair, we expect the semantic contents of them to be well-matched.
Yet it must be a disparity between a paired text and video.
To make full of the differentiated part of information and secure a comprehensive view of multimodal semantics, we consider a combined view of these two modalities.
Technically, with the above paired TSG and DSG at hand, we can unify them by creating a type of \textbf{cross-modal coreference edges}, $e^C_{v^T_i\leftrightarrow v^D_j}$, via which the objects in TSG link to the correspondence objects in DSG.
Specifically, we match the semantic similarities between any pair of text and image objects from TSG and DSG, respectively, via CLIP encoder \cite{RadfordKHRGASAM21}.
We measure the text label ($c^T_i$) against both the visual representation ($f^D_j$) and the visual object label ($c^D_j$). 
\begin{gather}
a^C(v^T_i,v^D_j) = \frac{1}{2} \left( \text{CLIP}( c^T_i , c^D_j ) + \text{CLIP}( c^T_i , f^D_j ) \right) \,,
\end{gather}
With a matching score $a^C(v^T_i,v^D_j)$ higher than $\gamma^C$ to be considered as the cross-modal co-referred nodes, we create a cross-modal coreference edge between the objects.
Note that the correspondence objects in DSG are only at the first occurred SG frame, which means that we only link the TSG nodes to the first potentially occurred nodes of DSG.
This results in an overall holistic SG (HSG) of both video and language, marked as $\bm{\mathcal{G}}^C$, as shown in Figure \ref{HSG}.
Formally, HSG can be given as:
\begin{gather}
\bm{\mathcal{G}}^C = \{\mathcal{G}^C_0,\mathcal{G}^C_1,\cdots,\mathcal{G}^C_t \, | \, \mathcal{E}^C_{v^T_i\leftrightarrow v^D_j} \} \,, 
\end{gather}
where HSG merges the single frame of TSG into the frame sequence of DSG as the first frame $\mathcal{G}^C_0$.
\begin{gather}
\mathcal{G}^C_0 = \mathcal{G}^{T} \,,  \,\,\,\,\,
\{\mathcal{G}^C_1,\cdots,\mathcal{G}^C_t \} = \{\mathcal{G}^D_1,\cdots,\mathcal{G}^D_t \} \,, 
\end{gather}
and the cross-modal coreference edges:
\begin{gather}
\mathcal{E}^{C}_{v^T\leftrightarrow v^D} = \{e^C_{v^T_i\leftrightarrow v^D_j}\}  \,.
\end{gather}

\section{Architecture of Finsta Framework}
\label{Framework Architecture}

We now present a fine-grained structural spatio-temporal alignment learning (dubbed Finsta) framework to encode the TSG, DSG and HSG representations that constituent the overall VLM system.
As shown in Figure \ref{framework1}(a), Finsta has a dual-stream-sum architecture.

\vspace{-2mm}
\subsection{Spatiality Encoding with Graph Transformer (GTrm)}
\label{GTrm}

First, for the TSG $G^{T}$, only the fine-grained spatiality of the scene should be taken care.
Thus we consider adopting a graph Transformer (GTrm) \cite{abs-2012-09699} to model the TSG.
Compared with the general graph neural networks \cite{marcheggiani-titov-2017-encoding,VelickovicCCRLB18}, GTrm advances in both the graph topology modeling and highly-parallel computation of Transformer architecture \cite{VaswaniSPUJGKP17}.
GTrm has $L$ stacked layers, where the update of representation $h_{i}^{l}$ of node $v_i$ in $l$-th layer is given by:
\begin{gather}
\overline{h^{l+1}_i} = O^l_{h,k}  \, ||_{k=1}^{H} \, \left( 
\sum_{j \in \mathcal{N}_i} w_{i,j}^{k,l} \, V^{k,l} \, h^{l}_i
\right) \,,  \\
\overline{h^{l+1}_i} = \text{Norm}(\overline{h^{l+1}_i}+ {h^{l}_i}) \,, \\
\label{h-base} h^{l+1}_i = \text{Norm}(\text{FFN}(\overline{h^{l+1}_i})+\overline{h^{l+1}_i}) \,,
\end{gather}
where $k$ denotes the attention head number, $O^l_k \in \mathbb{R}^{d\times d}$ is the attention head representation, $||$ is the concatenation.
The concatenation is followed by the Feed-Forward layer (FFN) and Add\&Norm layer with residual connection.
And $w_{i,j}^{k,l}$ is by the $k$-th self-attention head:
\begin{equation}
w_{i,j}^{k,l} = \text{Softmax}_j \left(
\frac{ Q^{k,l} \cdot K^{k,l} }{ \sqrt{d_{k_1}} }
\right) \cdot E^{k,l}  \,,
\end{equation}
where $E^{k,l}\in \mathbb{R}^{d\times d}$=${W}_{E}\{e_{i,j}\}$ is the embedding of edge $e^T_{i,j}$ in TSG.
Here all $K, Q, V \in \mathbb{R}^{d\times d}$ derived from the node representation via Eq \eqref{h-base} of last layer:
\begin{gather}\label{KQV}
K^{k,l}={W}_{K}\{h_{j}^{l}\} \,, \,
Q^{k,l}={W}_{Q}\{h_{j}^{l}\} \,, \,
V^{k,l}={W}_{V}\{h_{j}^{l}\} \,.
\end{gather}
The initial node representation $h_{i}^{0}$ is the embedding of the textual label of TSG node.
We gather all $h^{l+1}_i$ into ${H}^{l+1}$.

For the graph edge representation, there is a similar process as for the node propagation:
\begin{gather}
\overline{e^{l+1}_{i,j}} = O^l_{e,k}  \, ||_{k=1}^{H} \, \left( w_{i,j}^{k,l} \right) \,,  \\
\overline{e^{l+1}_{i,j}} = \text{Norm}(\overline{e^{l+1}_{i,j}}+ {e^{l}_{i,j}}) \,, \\
\label{e-base} e^{l+1}_{i,j} = \text{Norm}(\text{FFN}(\overline{e^{l+1}_{i,j}})+ \overline{e^{l}_{i,j}}) \,.
\end{gather}
By gather all $e^{l+1}_{i,j}$, we obtain the resulting $\bm{E}^{l+1}$.

\subsection{Spatiality-Temporality Encoding with Recurrent Graph Transformer (R-GTrm)}
\label{R-GTrm}

DSG is characterized by temporal dynamics, compared with the single-frame TSG.
Thus, we devise a novel recurrent graph Transformer (R-GTrm), for which we draw the main inspiration from the recurrent networks \cite{schuster1997bidirectional}.
As shown in Figure \ref{framework1}(b), built upon the GTrm propagation, the dynamics are additionally modeled in R-GTrm through the temporal coreference edges of nodes within DSG ($\bm{G}^D=\{G^D_1,\cdots,G^D_t\}$), essentially modeling the tracking of objects over time:
\begin{gather}
\overline{h^{l+1,t}_i} = O^{l,t}_{h,k}  \, ||_{k=1}^{H} \, \left( 
\sum_{j \in \mathcal{N}_i} w_{i,j}^{k,l,t} \, V^{k,l,t} \, h^{l,t}_i
\right) \,,  \\
\overline{h^{l+1,t}_i} = \text{Norm}(\overline{h^{l+1,t}_i}+ {h^{l,t}_i}) \,, \\
\label{h-base-recur} h^{l+1,t}_i = \text{Norm}(\text{FFN}(\overline{h^{l+1,t}_i})+\overline{h^{l+1,t}_i}) \,,
\end{gather}
Each attention head at $t$-th time-frame in DSG goes by
\begin{equation}
w_{i,j}^{k,l,t} = \text{Softmax}_j \left(
\frac{ \hat{Q}^{k,l,t} \cdot K^{k,l,t} }{ \sqrt{d_{k_2}} }
\right) \cdot E^{k,l,t}  \,.
\end{equation}
Here $E^{k,l,t}\in \mathbb{R}^{d\times d}$=${W}_{E}\{e_{i,j}^t\}$ is the embedding of edge $e^{D,t}_{i,j}$ of time-step $t$ in DSG.
Same as in Eq \eqref{KQV}, $K, Q, V$ are all derived from the node representation of corresponding frames via Eq \eqref{h-base-recur}.
The initial node representation $h_{i}^{0,t}$ is the concatenation of 1) the exported object neural representation $f_i^t$, 2) the embedding of the node label $c_i^t$ and 3) the embedding of the raw frame timestamp $\tau_i^t$.
Compared with GTrm, the update of R-GTrm representation $h_{i}^{l,t}$ of node $v_i$ at time-step $t$ further fuses the feature of prior frame $t$-$1$, via a automatic gate $\eta_q^t$:
\begin{gather}
\label{gating-1} \hat{Q}^{k,l,t} = (1-\eta_q^t) \cdot Q^{k,l,t-1} + \eta_q^t \cdot Q^{k,l,t} \,,  \\
\eta_q^t = \text{Sigma} ({W}_q  \cdot Q^{k,l,t} \cdot K^{k,l,t} ) \,.
\end{gather}
With $\hat{Q}^{k,l,t}$, we perform the same follow-up propagation.
The graph edge propagation shares the same process as for GTrm encoding TSG.
\begin{gather}
\overline{e^{l+1,t}_{i,j}} = O^l_{e,k,t}  \, ||_{k=1}^{H} \, ( w_{i,j}^{k,l,t} ) \,,  \\
\overline{e^{l+1,t}_{i,j}} = \text{Norm}(\overline{e^{l+1,t}_{i,j}}+ {e^{l,t}_{i,j}}) \,, \\
\label{e-base-recur} e^{l+1,t}_{i,j} = \text{Norm}(\text{FFN}(\overline{e^{l+1,t}_{i,j}})+ \overline{e^{l,t}_{i,j}}) \,.
\end{gather}
By gather all $e^{l+1,t}_{i,j}$, we obtain the resulting ${E}^{l+1,t}$.

As HSG has the same temporal property as DSG, we thus use another R-GTrm to encode HSG, i.e., $\bm{G}^C$:
\begin{equation}
\{H^C_0,H^C_1,\cdots H^C_t\} = \text{R-GTrm}^{C} ( \{G^C_0,G^C_1,\cdots,G^C_t\} )  \,.
\end{equation}
For simplicity, we denote the final TSG node feature matrix from GTrm as $H^T$=$\{h^T_1,\cdots,h^T_{i}\}$,
the matrix of DSG from R-GTrm as $\bm{H}^D$=$\{H^D_1,\cdots H^D_t\}$=$\{h^D_{1,1},\cdots,h^D_{t,i}\}$ ($t$ denotes the frame dimension),
and the one of HSG as $\bm{H}^C$=$\{H^C_0,H^C_1,\cdots H^C_t\}$.
Also, we use the resulting node representations of TSG and DSG from GTrm and R-GTrm (i.e., the last $L$-th layer), respectively, to initialize the node representations of HSG:
\begin{gather}
\label{HSG-initiation} \{H^{C,0}_0,H^{C,0}_1,\cdots H^{C,0}_t\} \gets \{ H^{T,L}  \| H^{D,L}_1,\cdots H^{D,L}_t\} \,.
\end{gather}

\begin{figure}[!t]
\begin{center}
\includegraphics[width=1\linewidth]{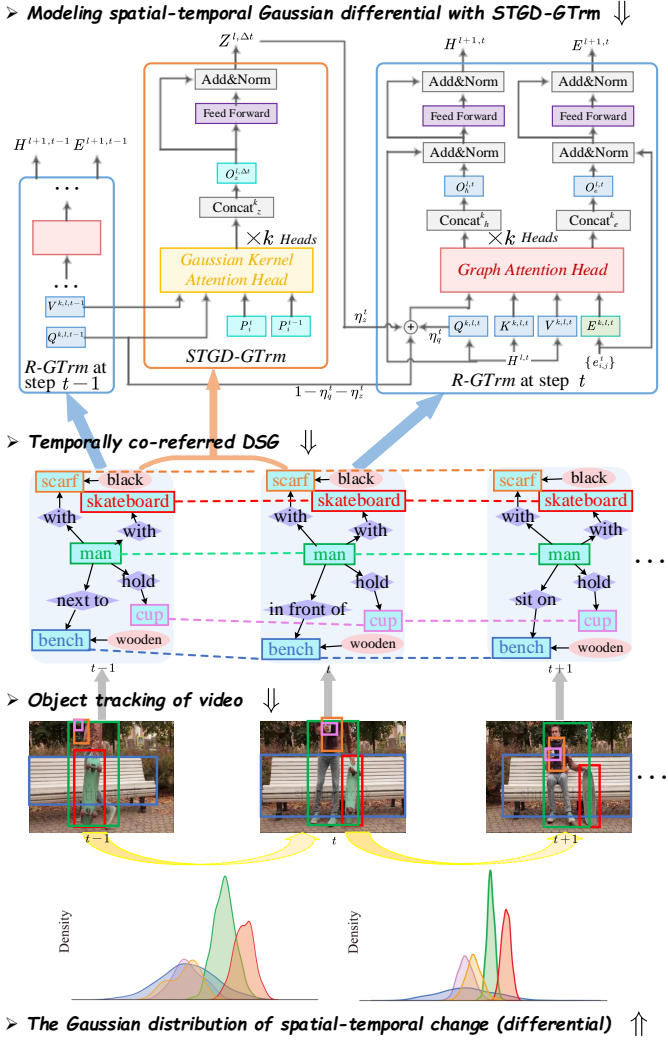}
\end{center}
\caption{
Illustration of the STGD-GTrm for modeling the spatio-temporal changes.
}
\vspace{-3mm}
\label{illus-dynamic-static}
\end{figure}

\subsection{Spatial-Temporal Gaussian Differential Graph Transformer (STGD-GTrm)}
\label{STGD-GTrm}

The above R-GTrm still fails to adequately perceive the changing spatial positions of objects.
A significant consequence here is the under-modeling of the distinction between stationary and moving objects. 
We emphasize that a key step in modeling video dynamics is to distinguish between objects in motion and those at rest (i.e., foreground vs. background).
To this end, based on the R-GTrm, we further design a spatial-temporal Gaussian differential graph Transformer (STGD-GTrm). 
The key idea is to enable the graph Transformer to perceive the changes in objects across spatial and temporal dimensions.
To illustrate this, we plot the strength (i.e., the distribution density, described by the following distribution kernel) of the changes of any objects between two consecutive frames (or keyframes) in Figure \ref{illus-dynamic-static}.
We see that such spatial-temporal change might intrinsically follow the Gaussian distribution, where those objects moving more clearly tend to have larger and sharper energy, while the stationary objects that move slowly tend to have lower strength.
Thus, we propose to model the spatio-temporal differential of the graph nodes of DSG or HSG along their trackers from the temporal coreference edges, with Gaussian distribution.
Technically, the spatial-temporal proximity of a node $v_i$ between the consecutive time from $t$ to $t^{'}$ is captured with a Gaussian kernel, depicted as follows:
\begin{equation} \label{kernel}
\kappa(v_i^{\Delta t}| \mathcal{N}_i) = \exp\left( -\frac{\| \mathcal{N}_i \| \cdot   \|p_{i}^t - p_{i}^{t-t^{'}}\|^2}{\sigma^z \cdot \sum_{j\ne i, j \in \mathcal{N}_i} \| p_{j}^t - p_{j}^{t-t^{'}}\|^2}  \right)  \,,
\end{equation}
where $\Delta t$=$t-t^{'}$, $\mathcal{N}_i$ denotes $v_i$ neighbor of adjacent nodes.
$p_{i}^t$ is the centroid of the object $v_i$ bounding box (marked as $\overline{b_{i}}$).
$\sigma^z$ is a spatial scale of the proximity.
By observing the spatial movement (position change) against the neighbors, it is also reasonable to distinguish those `\emph{false positive}' moving objects caused by the camera moving.

As shown in Figure \ref{illus-dynamic-static}, during the propagation of R-GTrm from $t$ to $t^{'}$ at $l$-th layer, an STGD-GTrm at this layer is inserted between these two R-GTrm frames.
With the $k$-th Gaussian kernel attention head $\kappa(v_i^{\Delta t})$ (denoted as $\kappa^{k,l,\Delta t}$ for unification), the self-attention encoder is given as:
\begin{equation}
z_{i}^{k,l,\Delta t} = \text{Softmax}_j \left(
\frac{ {Q}^{k,l,t^{'}} \cdot \kappa^{k,l,\Delta t} }{ \sqrt{d_{k_3}} }
\right) \cdot V^{k,l,t^{'}}  \,.
\end{equation}
With $k$ attention heads, we construct the resulting representation via concatenation, followed by the FFN transformation and residual connection, the same operation as in R-GTrm:
\begin{gather}
\overline{Z^{l,\Delta t}} = O^{l,\Delta t}_{z,k}  \, ||_{k=1}^{H} \, \left( 
\sum_{i}  \, z_{i}^{k,l,\Delta t}
\right) \,,  \\
Z^{l,\Delta t} = \text{Norm}(\text{FFN}(\overline{Z^{l,\Delta t}})+\overline{Z^{l,\Delta t}}) \,.
\end{gather}
With $Z^{l,\Delta t}$, we then retrofit the R-GTrm gating mechanism in Eq \eqref{gating-1} by fusing both the query feature of prior frame $t$-$1$ and this STGD-GTrm feature:
\begin{gather}
\label{gating-2} \hat{Q}^{k,l,t} = (1-\eta_q^t-\eta_z^t) \cdot Q^{k,l,t-1} + \eta_q^t \cdot Q^{k,l,t} + \eta_z^t \cdot Z^{l,\Delta t}\,,  \\
\eta_q^t = \text{Sigma} ({W}_q  \cdot Q^{k,l,t} \cdot K^{k,l,t} ) \,, \\
\eta_z^t = \text{Sigma} ({W}_z  \cdot Z^{l,\Delta t} \cdot K^{k,l,t} ) \,.
\end{gather}
With the new $\hat{Q}^{k,l,t}$, the following calculations in R-GTrm are carried out.
In this way, the system learns to better capture the spatiality-temporality changes of objects, and also is able to automatically recognize those nodes of \emph{static node} and \emph{dynamic node} during the graph representation learning.

\vspace{-2mm}
\section{Video-Language Representation Learning}
\label{Finsta-learning}

With the fine-grained structural features learned via the Finsta framework, we now perform representation learning, through which we enhance the VL representations of existing host VLMs.
In what follows, we first elaborate on cross-modal alignment learning.
Then we introduce how to apply our Finsta to an existing VLM.

\vspace{-2mm}
\subsection{Fine-grained Structural Spatio-Temporal Alignment Learning}
\label{Fine-grained Video-Language Alignment Learning}

We divide the VL alignment learning into the spatiality and temporality perspectives, where the former focuses on the fine-grained static object-level semantic matching, while the latter concentrates on the fine-grained dynamic motion-level semantic matching.
These two learning processes are carried out between the DSG and TSG encoding modules.\footnote{Note that, although the learning happens only between DSG and TSG encoders, the learned features will further propagate into the HSG encoder via the subsequent feature injection and initiation, cf. Eq \eqref{HSG-initiation}.}
Also we note that both two fine-grained alignment learning methods are automatically carried out unsupervisedly, i.e., without relying on external annotations or human interference.

\vspace{1mm}
\noindent\textbf{1) High-order Object-centered Spatial Contrasting (OSC).}
Our idea is to encourage the object nodes in TSG to find their correct correspondences in the DSG.
We adopt the contrastive learning \cite{He0WXG20} to pull semantically identical node pairs together, and push apart those different.
The fine-grained VL modeling can be carried out over single objects of texts and videos within the TSG and DSG.
However, we consider a more informative manner; we perform the matching of a high-order region that is centered on any objects.
Intuitively, a textual object and a visual object should be treated more similarly when the object pair as well as their modifying contexts (i.e., specific attributes and even relational neighbor objects) are all matched. 
Supplementary Material Section 1 illustrates the high-order neighbor modeling mechanism.
For a TSG object $v^T_i$, we traverse its $n$-th order (e.g., $1$st, $2$nd, or $3$rd) neighbors $\mathcal{N}^T_i$.
And then we obtain the region representation $\overline{h}^T_i$ via pooling operation.
Likewise, for a DSG object $v^D_{t,j}$, we also obtain the $n$-order neighbor representation $\overline{h}^D_{t,j}$.
We then measure the bipartite similarity between these two region representations, and then produce the learning target:
\begin{gather}\label{OSC-loss}
S^o_{i,t,j} = \frac{ (\overline{h}^{T}_i)^T \cdot \overline{h}^{D}_{t,j} }{ || \overline{h}^{T}_i ||  \, || \overline{h}^{D}_{t,j} ||  }   \,, \\
\mathcal{L}_{\text{\scriptsize OSC}}  = -  \sum_{i\in G^{T} ,\, j^{\ast}\in G^{D}_t }  \sum_{t}  \log 
\frac{ \exp(S^o_{i,t,j^{\ast}} /\tau^o ) }{\mathcal{Z}}  \,,
\end{gather}
where $\tau^o>$0 is an annealing factor.
We also define a threshold $\rho^o$ to decide the matching confidence, i.e., pairs with $S^o_{i,t,j}>\rho^o$ are considered aligned.
$j^{\ast}$ represents a positive DSG region with $i$ in TSG, i.e., $S^o_{i,t,j^{\ast}}$>$\rho^o$.

\begin{figure}[!t]
\centering
\includegraphics[width=0.82\linewidth]{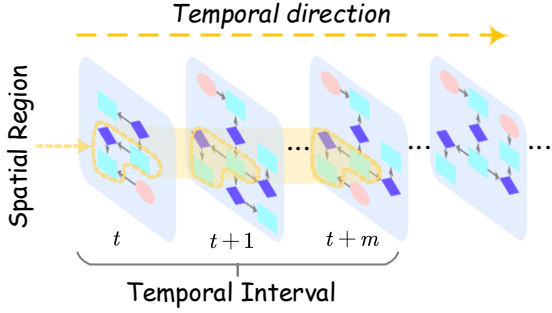}
\vspace{-1.5mm}
\caption{
The predicate-centered temporal contrasting mechanism, where we extract the spatial region and temporal interval for the predicate-centered temporal alignment.
}
\label{fig-PTC}
\vspace{-3mm}
\end{figure}

\vspace{1mm}
\noindent\textbf{2) High-order Predicate-centered Temporal Contrasting (PTC).}
Modeling only the spatiality is not enough, which motivates
the predicate-oriented dynamic semantics.
The predicate-centered temporal alignment has a similar formulation with OSC.
The aim is to find the correspondence between the textual predicates in TSG and the dynamic motions in DSG.
Slightly different from the OSC learning, we take a predicate-centered temporal contrasting learning.
Our targets are the dynamic relation nodes (i.e., predicates) in TSG, and centered on a predicate node $v^T_i$.
Likewise, we first find its $n$-order neighbor spatial region (the representation is marked as $\hat{h}^T_i$) within TSG.
We use the same method to find its $n$-order neighbor spatial region (the representation is marked as $\hat{h}^T_i$) within TSG.

Then, in the DSG, we also find such $n$-order region for each predicate node $v^D_{t,j}$,
and further slice the DSG sequence with a temporal interval, $t:t+m$, i.e., starting from $t$-th frame and ending at ($t+m$)-th frame
in the DSG sequence.
We take the pooled representation over the region features ($\hat{h}^D_{t:t+m,j}$) as the candidate counterpart of DSG.
This is illustrated in Figure \ref{fig-PTC}.
Thereafter, we perform PTC learning:
\begin{gather}\label{PTC-loss}
S^p_{i,t:t+m,j} = \frac{ (\hat{h}^{T}_i)^T \cdot \hat{h}^{D}_{t:t+m,j} }{ || \hat{h}^{T}_i ||  \, || \hat{h}^{D}_{t:t+m,j} ||  }   \,, \\
\mathcal{L}_{\text{\scriptsize PTC}}  = - \sum_{i\in G^{T} ,\, j^{\ast}\in G^{D}_t }  \sum_{t}  \sum_{m}  \log 
\frac{ \exp(S^p_{i,t:t+m,j^{\ast}} /\tau^p ) }{\mathcal{Z}}  \,,
\end{gather}
where $\tau^p$ is the annealing factor, and $\rho^p$ is the threshold for PTC.
Such `\emph{textual predicate}'-`\emph{visual object tracking}' alignment vividly simulates the temporal dynamics of two modalities.

\subsection{Representation Transfer Learning}
\label{Representation Interference Learning}

Via the above alignment learning, the TSG and DSG representations of text and video can be well matched, and expected to better facilitate downstream VL tasks.
However, directly applying Finsta as VLM for usage can be problematic, because our system relies heavily on the SG annotations, while parsing SG labels for all potential incoming data will inevitably introduce noises and result in low-efficient applications.
Meanwhile, training a VLM with Finsta from scratch can be much resource-costing (i.e., with 100m VL pairs), and parsing such a large number of SG annotations is impractical.
To this end, we consider designing our Finsta as a plug-and-play module, and inject the well-aligned VL feature representations into a host VLM.
Based on any existing VLM with the similar dual-stream-sum architecture, with a warm-starting, we can incrementally perform the alignment more efficiently.

\begin{figure}[!t]
\centering
\includegraphics[width=0.82\linewidth]{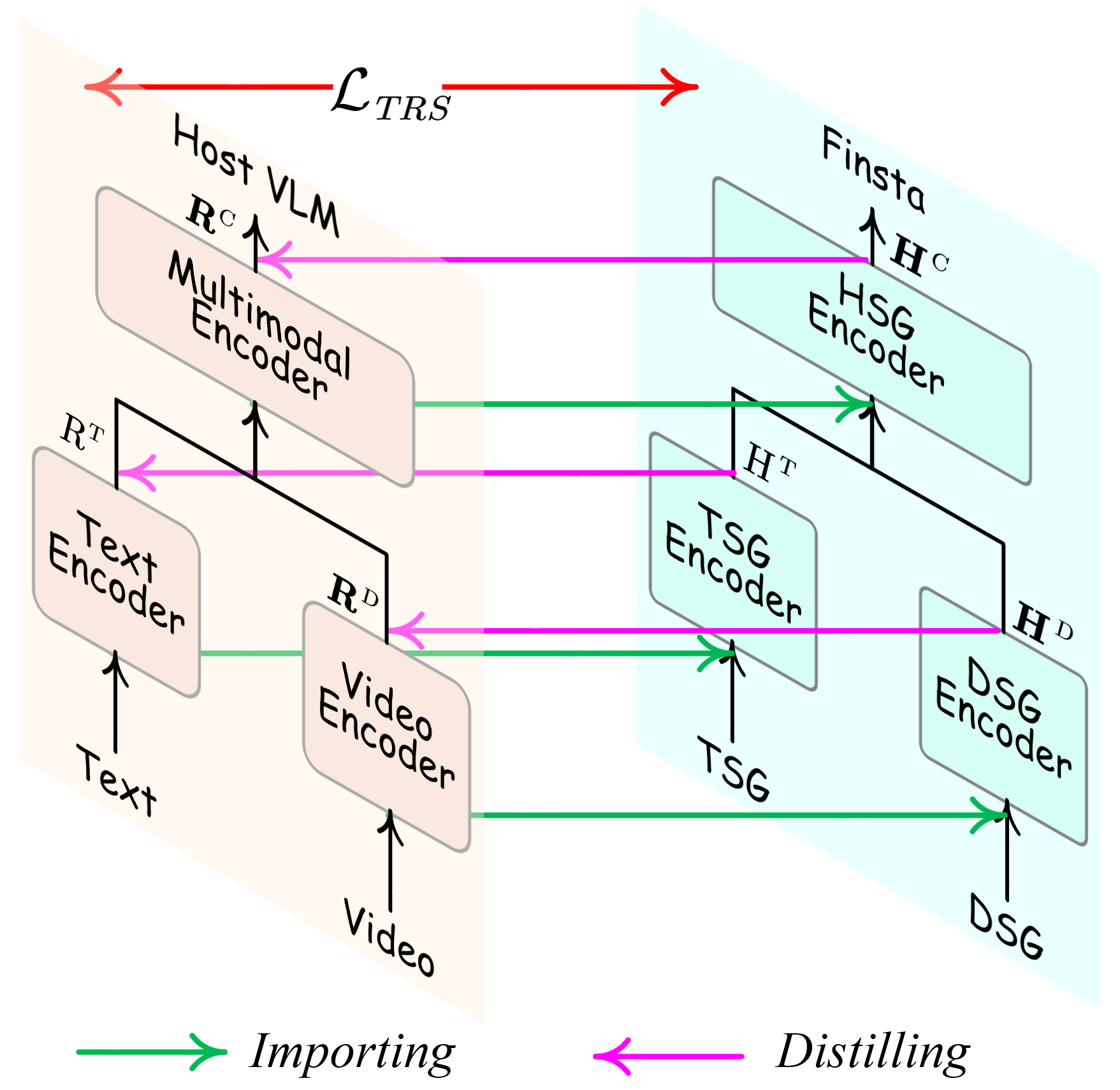}
\caption{
A stereoscopic illustration of registering the representations from Finsta system into a host VLM.
}
\label{register}
\vspace{-4mm}
\end{figure}

Technically, we register Finsta to a host VLM with the knowledge distillation (KD) technique \cite{HintonVD15,FurlanelloLTIA18}.
Figure \ref{register} illustrates the mechanism.
Before Finsta propagates the messages, we first import the first-layer representations of the text encoder, video encoder and multimodal encoder of host VLM respectively, into Finsta's three key modules as the initial feature representations of various SG modeling, which are seen as well-aligned visual-language embeddings.
We use ${R}^T$, $\bm{R}^D$ and $\bm{R}^C$ to denote the representations of text, video and multimodal encoder in host VLM, respectively.
Specifically, the text/video/multimodal representations of the first-layer host VLM, $\bm{R}^{T/D/C,l_1}$, are copied to Finsta as the initial input feature embeddings, i.e., $\bm{H}^{T/D/C,l_0} \gets \bm{X} + \bm{R}^{T/D/C,l_1}$, where $X$ is the node embedding of the input TSG/VSG/HSG in Finsta.
By injecting the well-initiated VL feature representations into Finsta, we can warm-start the following post-training of Finsta with host VLM.

Afterward, we distill the features of Finsta into host VLM.
The Finsta encoder performs propagations over the SG data, and finally obtains the resulting well-aligned spatio-temporal VL feature $\bm{H}^{T/D/C}$ in the last layer.
Next, we distill them from Finsta to the host VLM via KD.
Such that, the host VLM encoding features learn to be similar to the ones as in Finsta, i.e., the fine-grained spatial-temporally aligned features.
\begin{equation} \label{KD}
\begin{aligned}
\mathcal{L}_{\text{\scriptsize TRS}}  = \sum_d (
&\lambda^T || {R}^T - H^T ||^2 + \\
&\lambda^D \sum_t  || \bm{R}_t^D - \bm{H}_t^D ||^2  + \\
&\lambda^C \sum_t  || \bm{R}_t^C - \bm{H}_t^C ||^2  ) \,,
\end{aligned}
\end{equation}
where $\lambda^{T/D/C}$ are the learning co-efficiencies. 
We note that the representation used for distillation is the overall instance-level representation, i.e., we take the output representation of the `[CLS]' token as the transfer targets. 
Also, the KD for feature injection only happens during the post-training phase.
During fine-tuning or inference for the downstream task and data, the host VLM is able to make better predictions alone without the attendance of Finsta.
This way, the SG annotations are only needed during the post-training stage.

\vspace{2mm}
\noindent\textbf{Remark.}
We expect the host VLMs to have the same architecture (i.e., text encoder, video encoder and cross-modal encoder) as Finsta to achieve the plug-and-play functionality. 
\emph{Exactly same architecture}, yet mostly necessary, is not a strict requirement.
While the `\emph{text-video-multimodal}' encoding architecture has been the standard paradigm in most of the existing VLMs \cite{abs-2111-12681}, there are also a number of VLMs that do not come with strictly the dual-stream-sum architecture \cite{xu-etal-2021-videoclip,lin2023video}.
Even if any of the three encoders are absent in the host VLM, Finsta can still work, by not distilling the Finsta features of the counterpart encoder to the absent encoder in the host VLM (i.e., in Eq \ref{KD}, remove any of the three). 
But in such case,
the efficacy of Finsta will be sacrificed to a certain extent, which we will analyze in the following experiment section \ref{Influence of VLM Module Existence}.

\subsection{Training of Overall System}
\label{Overall Training}

The training of the overall framework takes a warm-up paradigm.
We first pre-train the Finsta alone on the text-video pairs with TSG and DSG annotations, with the alignment learning ($\mathcal{L}_{\text{\scriptsize PTC}}$ and $\mathcal{L}_{\text{\scriptsize OSC}}$).
When the Finsta tends to converge, we then perform the knowledge distillation, and inject the Finsta representations into the host VLM as described above.
The joint training involves three learning objectives: 
$\mathcal{L}_{\text{\scriptsize PTC}}$, $\mathcal{L}_{\text{\scriptsize OSC}}$ and $\mathcal{L}_{\text{\scriptsize TRS}}$.
By the way, there are also the standard VL learning objectives in the host VLM, such as the masked language modeling $\mathcal{L}_{\text{\scriptsize MLM}}$ \cite{SunXS00F22}, and the overall coarse-grained video-text alignment learning $\mathcal{L}_{\text{\scriptsize VLA}}$ \cite{XueHZS00FG22}.
We can summarize all the learning objectives together:
\begin{equation}
\begin{aligned}
\mathcal{L}  = 
\lambda^{OSC} \mathcal{L}_{\text{\scriptsize OSC}} +  
\lambda^{PTC} \mathcal{L}_{\text{\scriptsize PTC}} + 
\lambda^{TRS} \mathcal{L}_{\text{\scriptsize TRS}} \\ + 
\lambda^{MLM} \mathcal{L}_{\text{\scriptsize MLM}} + 
\lambda^{VLA} \mathcal{L}_{\text{\scriptsize VLA}} \,,
\end{aligned}
\end{equation}
where $\lambda^{*}$ are the co-efficiencies that dynamically change by linearly learning scheduler \cite{huang-etal-2020}.

\section{Experiments and Main Results}
\label{Experimental Settings}

\subsection{Experimental Settings}

\noindent\textbf{1) Video-Language Understanding Tasking}
There has been a series of representative VL modeling tasks, which in our experiments we divide into four groups:
video-to-text transformation (e.g., Video Action Recognition, Video Captioning), text-to-video transformation (e.g., Video-Text Retrieval), VL collaboration (e.g., Video Question Answering) and the more challenging scenario, long-form VL understanding (e.g., Long-Form Video Question Answering, Video-Paragraph Retrieval).
For each task, we use the representative datasets, and measure the performance with metrics by following the common practice.
In Supplementary Material Section 2 we give a detailed description of all tasks with respect to the task definitions, datasets and metrics.

\vspace{1mm}
\noindent\textbf{2) Implement Details}. 
Our Finsta takes a 12-layer GTrm for TSG encoding and 12-layer R-GTrm \& STGD-GTrm for DSG encoding ($L$=12).
The HSG R-GTrm \& STGD-GTrm encoders is a 6-layer version.
All the attention head number is 8 ($k$=8).
We set all dimensions as 768 in our system. 
The hyper-parameters during post-training are set as follows, which help achieve the best effects.
Initial annealing factor $\tau^o$ and $\tau^p$ are all set as 0.8.
The co-efficiencies in Eq \ref{KD} are set as [$\lambda^{T}$=0.2,$\lambda^{D}$=0.5,$\lambda^{C}$=0.3] for video-to-text transformation tasks, and [$\lambda^{T}$=0.3,$\lambda^{D}$=0.35,$\lambda^{C}$=0.35] for VL co-comprehension tasks, and [$\lambda^{T}$=0.35,$\lambda^{D}$=0.3,$\lambda^{C}$=0.35] for text-to-video transformation tasks.
The initial weights are set as: $\lambda^{OSC}$=0.5, $\lambda^{PTC}$=0.5, both of them will be linearly decreased from 0.5 to 0.15 along the training.
And $\lambda^{TRS}$=0.2 gradually increases to 0.5.
$\lambda^{VLA}$=0.3 and $\lambda^{MLM}$=0.3 are kept unchanged.
The threshold for building the temporal coreference edges, $\gamma^D$ is set as 0.6; and for cross-modal coreference edges, $\gamma^C$ is 0.9.
The $n$ in $n$-order neighboring calculation for alignment learning is set as 3 for OSC and 4 for PTC.
The alignment confidence threshold value $\rho^o$ is set as 0.7 for OSC, and $\rho^p$ is set as 0.6 for PTC.

\vspace{1mm}
\noindent\textbf{3) Baselines and Backbone VLMs}. 
We compare with strong-performing baselines of different benchmarks.
We consider the existing state-of-the-art language models as our backbones,
including 10 VLMs and 3 LVLMs.
We adopt VLMs which may either have 1) dual-stream-sum architectures with three `\emph{text-video-multimodal}' encoders, such as HDVILA \cite{XueHZS00FG22}, Clover \cite{abs-2207-07885}, LFVILA \cite{SunXS00F22}, or 2) with certain encoder(s) being absent, such as VideoCLIP \cite{xu-etal-2021-videoclip} \& CLIP4Clip \cite{LuoJZCLDL22} missing the cross-modal encoder, and All-in-one \cite{wang2023all} only having one multimodal encoder.
Different (L)VLMs are pre-trained on different amounts of corpus, and with different volumes of parameters.
We elaborate all the VLMs' architectures, parameter sizes and pertaining data in Supplementary Material Section 3.1.

\vspace{1mm}
\noindent\textbf{4) Post-Training Details}.
In the post-training, we first tune the Finsta alone for 2 epochs as warming-up, using a batch size of 300. 
We use AdamW optimizer with a weight decay 5e-3 and betas (0.9, 0.98).
The learning rate is first warmed-up by 1 epoch to 5e-3 and then decays.
All trainings are conducted on 16 NVIDIA A100 GPUs.
For the post-training of Finsta-VLMs on the normal (short) form videos, we use a total of 50K VL pairs, with 25K sampled from WebVid-2.5M \cite{BainNVZ21} and 25K sampled from HD-VILA-100M \cite{XueHZS00FG22}.
For the post-training of Finsta-LFVILA\footnote{We denote the Finsta-LFVILA post-trained on the normal (short) form videos as \emph{S-Vid}, and on the long-form data as \emph{L-Vid}.}, we further consider the use of long-form data as what LFVILA has been trained, i.e., with 30K VL pairs sampled from LF-VILA-8M \cite{SunXS00F22}, where the avg. video duration is 100.2 sec. and the avg. text length is 307.9 tokens.
Supplementary Material Section 3.2 extends the details of post-training datasets.

\noindent\textbf{5) Scene Graph Parsing}. 
For the TSG annotations, we mainly follow the prior practice of SG applications.
We also perform filtering to remove objects, relations, and attributes that appear less than 5 times in all the parsed scene graphs. 
After such filtering, we obtain 7,021 objects, 2,256 relations, and 4,895 attributes in TSGs. 
For each video, we flexibly extract 10-50 keyframes while preserving their order.
The key process of DSG parsing has been elaborated in Section \ref{constrction-DSG}.

\vspace{1mm}
\noindent\textbf{6) End-Task Fine-tuning Details}.
For the input videos of the host VLM, we resize and center crop the video frames into 256$\times$256 to split into patches with size 16$\times$16, getting H=W=16. 
Then, the joint training with host VLM is with epochs from 5 to 20, using a batch size in [100,150,200], flexibly dependent on the task and dataset used.
For different downstream tasks, we mainly keep the same configuration as described in the above implementation details, with only few places are further tuned.
The scores of different models from our implementations are averaged five runs with random seeds, and the results of other baselines are copied from the raw papers (for which we will mark their citations).

\begin{table}[t]
\centering
\caption{Video Action Recognition results (Acc. on Top-1 \& Top-5) on two datasets. 
The best results are in \textcolor{mblue}{\textbf{blue}}, while the existing state-of-the-art results are \underline{underlined}.
\textcolor{mred}{Red} scores are the improvement of \textbf{\color{finsta-color} Finsta} over the backbone VLMs.
}
\fontsize{9}{10}\selectfont
\setlength{\tabcolsep}{1.3mm}
\begin{tabular}{lllll}
\hline
\multicolumn{1}{c}{\multirow{2}{*}{\bf Method}}&   \multicolumn{2}{c}{\bf K400 \cite{KayCSZHVVGBNSZ17}} & \multicolumn{2}{c}{\bf SSV2 \cite{GoyalKMMWKHFYMH17}} \\
\cmidrule(r){2-3}\cmidrule(r){4-5}
 & \bf Top-1	& \bf Top-5  & \bf Top-1& \bf Top-5 \\   
\midrule
TimeSformer \cite{BertasiusWT21}&	78.0&	93.7 & 59.5& -\\   
Frozen \cite{BainNVZ21}&	78.5& 94.1&	61.6& 85.7\\   
OmniVL \cite{Wang0WLZZX0JY22}&	79.1& 94.5&	62.5& 86.2\\  
ViViT \cite{arnab2021vivit}  &84.9  &95.8 &65.4  & 89.8\\
Swin \cite{liu2022video}  &84.9  & 96.7 &69.6  & 92.7\\
UniFormer-V2 \cite{li2022uniformerv2}& 85.4& 97.0&	73.0& 94.5\\
VideoMAE \cite{tong2022videomae}  & 87.4& 97.6& 75.4& 95.2\\

\hline

HDVILA  &	78.6& 94.0 &	61.3 & 86.2\\   
\rowcolor{blue} \textbf{\color{finsta-color} Finsta}-HDVILA & 83.4\textcolor{mred}{\scriptsize{\textbf{ +4.8}}} & 97.6\textcolor{mred}{\scriptsize{\textbf{ +3.6}}} & 65.2\textcolor{mred}{\scriptsize{\textbf{ +3.9}}} & 90.3\textcolor{mred}{\scriptsize{\textbf{ +4.1}}} \\

Clover 	&78.8& 95.3 &	62.3 & 86.9\\   
\rowcolor{blue} \textbf{\color{finsta-color} Finsta}-Clover & 81.2\textcolor{mred}{\scriptsize{\textbf{ +2.4}}} & 97.1\textcolor{mred}{\scriptsize{\textbf{ +1.8}}} & 64.1\textcolor{mred}{\scriptsize{\textbf{ +1.8}}} & 91.4\textcolor{mred}{\scriptsize{\textbf{ +4.5}}}\\

InternVideo    & \underline{91.1}&	\underline{98.9} & \underline{77.2}& \underline{95.4}\\
\rowcolor{blue} \textbf{\color{finsta-color} Finsta}-InternVideo   & \color{mblue} \bf  93.7\textcolor{mred}{\scriptsize{\textbf{ +2.6}}}& \color{mblue} \bf 99.2\textcolor{mred}{\scriptsize{\textbf{ +0.3}}} & \color{mblue} \bf 80.5\textcolor{mred}{\scriptsize{\textbf{ +3.3}}}& \color{mblue} \bf 96.7\textcolor{mred}{\scriptsize{\textbf{ +1.3}}} \\

\hdashline
Video-LLaMA   & 86.9&	97.1& 74.7&	93.6 \\
\rowcolor{blue} \textbf{\color{finsta-color} Finsta}-Video-LLaMA  & 91.2\textcolor{mred}{\scriptsize{\textbf{ +4.3}}}& 98.0\textcolor{mred}{\scriptsize{\textbf{ +0.9}}}& 76.8\textcolor{mred}{\scriptsize{\textbf{ +2.1}}}& 94.5\textcolor{mred}{\scriptsize{\textbf{ +0.9}}} \\

Video-LLaVA   & 88.3&	97.8& 75.8&	94.0 \\
\rowcolor{blue} \textbf{\color{finsta-color} Finsta}-Video-LLaVA  & 92.4\textcolor{mred}{\scriptsize{\textbf{ +4.2}}}& 98.8\textcolor{mred}{\scriptsize{\textbf{ +1.0}}}& 77.0\textcolor{mred}{\scriptsize{\textbf{ +1.2}}}& 95.6\textcolor{mred}{\scriptsize{\textbf{ +1.6}}} \\

\hline
\end{tabular}
\vspace{-3mm}
\label{tab:VAR}
\end{table}

\begin{table}[t]
\centering
\caption{Video Captioning results on two datasets.
M: METEOR; B@4: BLEU@4.
}
\vspace{-2mm}
\fontsize{9}{10}\selectfont
\setlength{\tabcolsep}{1.4mm}
\begin{tabular}{lllll}
\hline
\multicolumn{1}{c}{\multirow{2}{*}{\bf Method}} & \multicolumn{2}{c}{\bf YouCook2 \cite{ZhouXC18}} & \multicolumn{2}{c}{\bf MSR-VTT \cite{XuMYR16}} \\
\cmidrule(r){2-3} \cmidrule(r){4-5}
& \bf M & \bf B@4 & \bf M & \bf B@4 \\
\midrule
VideoBERT \cite{SunMV0S19} & 11.0 & 4.1 & - & - \\
OmniVL \cite{Wang0WLZZX0JY22} & 14.8 & 8.7 & - & - \\
IcoCap \cite{liang2023icocap} & - & - & 30.3 & 46.1 \\
HiTeA \cite{ye2023hitea} & - & - & 30.7 & 49.2 \\
VLAB \cite{he2023vlab} & - & - & 33.4 & 54.6 \\

\hline

UniVL [6] & \underline{22.4} & \underline{17.3} & 29.7 & 45.7 \\
\rowcolor{blue} \textbf{\color{finsta-color} Finsta}-UniVL & \color{mblue} \bf 23.6\textcolor{mred}{\scriptsize{\textbf{ +1.2}}} & \color{mblue} \bf 18.4\textcolor{mred}{\scriptsize{\textbf{ +1.1}}} & 33.4\textcolor{mred}{\scriptsize{\textbf{ +3.7}}} & 49.0\textcolor{mred}{\scriptsize{\textbf{ +3.3}}} \\

HDVILA & 13.5 & 8.2 & 32.4 & 46.0 \\
\rowcolor{blue} \textbf{\color{finsta-color} Finsta}-HDVILA & 18.8\textcolor{mred}{\scriptsize{\textbf{ +5.3}}} & 12.7\textcolor{mred}{\scriptsize{\textbf{ +4.5}}} & 36.9\textcolor{mred}{\scriptsize{\textbf{ +4.5}}} & 48.6\textcolor{mred}{\scriptsize{\textbf{ +2.6}}} \\

Clover & 14.2 & 9.0 & \underline{34.1} & 47.5 \\
\rowcolor{blue} \textbf{\color{finsta-color} Finsta}-Clover & 18.6\textcolor{mred}{\scriptsize{\textbf{ +4.4}}} & 12.5\textcolor{mred}{\scriptsize{\textbf{ +3.5}}} & \color{mblue} \bf38.8\textcolor{mred}{\scriptsize{\textbf{ +4.7}}} & 49.3\textcolor{mred}{\scriptsize{\textbf{ +1.8}}} \\

All-in-one & 12.0 & 7.2 & 32.7 & 48.3 \\
\rowcolor{blue} \textbf{\color{finsta-color} Finsta}-All-in-one & 13.1\textcolor{mred}{\scriptsize{\textbf{ +1.1}}} & 7.6\textcolor{mred}{\scriptsize{\textbf{ +0.4}}} & 34.2\textcolor{mred}{\scriptsize{\textbf{ +1.5}}} & 50.0\textcolor{mred}{\scriptsize{\textbf{ +1.7}}} \\

\hdashline

Video-LLaMA & 18.5 & 12.7 & 30.7 & 56.3 \\
\rowcolor{blue} \textbf{\color{finsta-color} Finsta}-Video-LLaMA & 21.3\textcolor{mred}{\scriptsize{\textbf{ +2.8}}} & 16.3\textcolor{mred}{\scriptsize{\textbf{ +3.6}}} & 33.0\textcolor{mred}{\scriptsize{\textbf{ +2.3}}} & 57.1\textcolor{mred}{\scriptsize{\textbf{ +0.8}}} \\

Video-LLaVA & 20.0 & 16.4 & 33.5 & \underline{57.0} \\
\rowcolor{blue} \textbf{\color{finsta-color} Finsta}-Video-LLaVA & 23.0\textcolor{mred}{\scriptsize{\textbf{ +3.0}}} & 17.0\textcolor{mred}{\scriptsize{\textbf{ +0.6}}} & 35.3\textcolor{mred}{\scriptsize{\textbf{ +1.8}}} & \color{mblue} \bf58.9\textcolor{mred}{\scriptsize{\textbf{ +1.9}}} \\

\hline
\end{tabular}
\label{tab:VC}
\end{table}

\begin{table}[t]
\centering
\caption{Video-Text Retrieval results on two datasets.}
\vspace{-2mm}
\fontsize{9}{10}\selectfont
\setlength{\tabcolsep}{1.0mm}
\begin{tabular}{lllll}
\hline
\multicolumn{1}{c}{\multirow{2}{*}{\bf Method}}&   \multicolumn{2}{c}{\bf LSMDC \cite{MaharajBRCP17}} & \multicolumn{2}{c}{\bf DiDeMo \cite{HendricksWSSDR17}} \\
\cmidrule(r){2-3}\cmidrule(r){4-5}
 & \bf R@1 & \bf R@5 & \bf R@1 & \bf R@5 \\
\midrule
OA-Trans \cite{WangGCY0SQS22} & 18.2 & 34.3 & 34.8 & 64.4 \\
ALPRO \cite{Li0LNH22} & - & - & 35.9 & 67.5 \\
Frozen \cite{BainNVZ21} & 15.0 & 30.8 & 31.0 & 59.8 \\
CAMoE \cite{abs-2109-04290} & 22.5 & 42.6 & 43.8 & 71.4 \\
UMT-L \cite{li2023unmasked} & \underline{43.0} & \underline{65.5} & 70.4 & \underline{90.1} \\

\hline

VIOLET & 16.1 & 36.6 & 32.6 & 62.8 \\
\rowcolor{blue} \textbf{\color{finsta-color} Finsta}-VIOLET & 20.4\textcolor{mred}{\scriptsize{\textbf{ +4.3}}} & 39.1\textcolor{mred}{\scriptsize{\textbf{ +2.5}}} & 37.8\textcolor{mred}{\scriptsize{\textbf{ +5.2}}} & 66.0\textcolor{mred}{\scriptsize{\textbf{ +3.2}}} \\

VideoCLIP & 18.6 & 36.0 & 32.8 & 63.0 \\
\rowcolor{blue} \textbf{\color{finsta-color} Finsta}-VideoCLIP & 21.8\textcolor{mred}{\scriptsize{\textbf{ +3.2}}} & 38.3\textcolor{mred}{\scriptsize{\textbf{ +2.3}}} & 35.5\textcolor{mred}{\scriptsize{\textbf{ +2.7}}} & 64.5\textcolor{mred}{\scriptsize{\textbf{ +1.5}}} \\

CLIP4Clip & 21.6 & 41.8 & 43.4 & 70.2 \\
\rowcolor{blue} \textbf{\color{finsta-color} Finsta}-CLIP4Clip & 22.9\textcolor{mred}{\scriptsize{\textbf{ +1.3}}} & 43.0\textcolor{mred}{\scriptsize{\textbf{ +1.2}}} & 45.2\textcolor{mred}{\scriptsize{\textbf{ +1.8}}} & 71.3\textcolor{mred}{\scriptsize{\textbf{ +1.1}}} \\

MCQ & 17.9 & 35.4 & 37.0 & 62.2 \\
\rowcolor{blue} \textbf{\color{finsta-color} Finsta}-MCQ & 22.8\textcolor{mred}{\scriptsize{\textbf{ +4.9}}} & 40.2\textcolor{mred}{\scriptsize{\textbf{ +4.8}}} & 41.7\textcolor{mred}{\scriptsize{\textbf{ +4.7}}} & 66.5\textcolor{mred}{\scriptsize{\textbf{ +4.3}}} \\
HDVILA & 17.4 & 34.1 & 28.8 & 57.4 \\
\rowcolor{blue} \textbf{\color{finsta-color} Finsta}-HDVILA & 25.3\textcolor{mred}{\scriptsize{\textbf{ +7.9}}} & 46.3\textcolor{mred}{\scriptsize{\textbf{ +12.2}}} & 41.3\textcolor{mred}{\scriptsize{\textbf{ +12.5}}} & 70.9\textcolor{mred}{\scriptsize{\textbf{ +13.5}}} \\

Clover & 24.8 & 44.0 & 50.1 & 76.7 \\
\rowcolor{blue} \textbf{\color{finsta-color} Finsta}-Clover & 28.9\textcolor{mred}{\scriptsize{\textbf{ +4.1}}} & 48.8\textcolor{mred}{\scriptsize{\textbf{ +4.8}}} & 56.0\textcolor{mred}{\scriptsize{\textbf{ +5.9}}} & 82.8\textcolor{mred}{\scriptsize{\textbf{ +6.1}}} \\

All-in-one & 20.5 & 38.8 & 32.7 & 61.4 \\
\rowcolor{blue} \textbf{\color{finsta-color} Finsta}-All-in-one & 22.0\textcolor{mred}{\scriptsize{\textbf{ +1.5}}} & 39.9\textcolor{mred}{\scriptsize{\textbf{ +1.1}}} & 33.2\textcolor{mred}{\scriptsize{\textbf{ +0.5}}} & 62.1\textcolor{mred}{\scriptsize{\textbf{ +0.7}}} \\

\hdashline

Video-LLaMA & 38.0 & 62.5 & 67.5 & 82.4 \\
\rowcolor{blue} \textbf{\color{finsta-color} Finsta}-Video-LLaMA & \color{mblue} \bf44.8\textcolor{mred}{\scriptsize{\textbf{ +6.8}}} & 66.7\textcolor{mred}{\scriptsize{\textbf{ +4.2}}} & 70.2\textcolor{mred}{\scriptsize{\textbf{ +2.7}}} & 85.7\textcolor{mred}{\scriptsize{\textbf{ +3.3}}} \\

Video-LLaVA & 40.6 & 61.5 & \underline{71.2} & {88.7} \\
\rowcolor{blue} \textbf{\color{finsta-color} Finsta}-Video-LLaVA & 43.5\textcolor{mred}{\scriptsize{\textbf{ +2.9}}} & \color{mblue} \bf67.2\textcolor{mred}{\scriptsize{\textbf{ +5.7}}} & \color{mblue} \bf73.6\textcolor{mred}{\scriptsize{\textbf{ +2.4}}} & \color{mblue} \bf90.3\textcolor{mred}{\scriptsize{\textbf{ +1.6}}} \\

\hline
\end{tabular}
\vspace{-3mm}
\label{tab:VTR}
\end{table}

\begin{table}[t]
\centering
\caption{Video Question Answering results (Acc.) on two datasets: MSR-VTT-QA \cite{XuZX0Z0Z17}, MSVD-QA \cite{XuZX0Z0Z17}, TGIF-Frame \cite{JangSYKK17}.
MC: Multiple-choice type;
OE: Open-ended type.
}
\vspace{-2mm}
\fontsize{9}{10}\selectfont
\setlength{\tabcolsep}{1.4mm}
\begin{tabular}{lllll}
\hline

\multicolumn{1}{c}{\multirow{2}{*}{\bf Method}}& \multicolumn{2}{c}{\bf MSR-VTT } & \multicolumn{1}{c}{\bf MSVD } & \multicolumn{1}{c}{\bf TGIF} \\
\cmidrule(r){2-3}\cmidrule(r){4-4}\cmidrule(r){5-5}
 & \bf MC & \bf OE & \bf OE &  \bf OE  \\   
\midrule

ClipBERT \cite{LeiLZGBB021} & 88.2 & 37.4 & 48.7  & 64.7  \\   
ALPRO \cite{Li0LNH22} & - & 42.1 & 45.9  & -  \\   
OmniVL \cite{Wang0WLZZX0JY22} & - & 44.1 & 51.0  &  - \\   
HiTeA \cite{ye2023hitea} & \underline{97.2} & 45.4 & 55.6  &  73.2 \\
VLAB \cite{he2023vlab}& - & \underline{49.6} & 61.0  &  \underline{79.0} \\ 

\hline

VIOLET & 91.9 & 43.9 & 47.9  & 65.3  \\   
\rowcolor{blue} \textbf{\color{finsta-color} Finsta}-VIOLET & 94.6\textcolor{mred}{\scriptsize{\textbf{ +2.7}}}& 47.4\textcolor{mred}{\scriptsize{\textbf{ +3.5}}}& 54.6\textcolor{mred}{\scriptsize{\textbf{ +6.7}}}& 69.8\textcolor{mred}{\scriptsize{\textbf{ +4.5}}} \\

VideoCLIP & 92.1 & 41.3 & 48.9  & 67.0 \\
\rowcolor{blue} \textbf{\color{finsta-color} Finsta}-VideoCLIP & 95.3\textcolor{mred}{\scriptsize{\textbf{ +3.2}}}& 45.8\textcolor{mred}{\scriptsize{\textbf{ +4.5}}}& 53.7\textcolor{mred}{\scriptsize{\textbf{ +4.8}}}& 69.2\textcolor{mred}{\scriptsize{\textbf{ +2.2}}} \\

HDVILA & 93.1 & 40.0 & 50.7  & 68.3  \\   
\rowcolor{blue} \textbf{\color{finsta-color} Finsta}-HDVILA & 96.3\textcolor{mred}{\scriptsize{\textbf{ +3.2}}}& 45.4\textcolor{mred}{\scriptsize{\textbf{ +5.4}}}& 53.3\textcolor{mred}{\scriptsize{\textbf{ +2.6}}}& 71.8\textcolor{mred}{\scriptsize{\textbf{ +3.5}}}  \\  

Clover & 95.0 & 42.5 & 51.1 &  71.6 \\   
\rowcolor{blue} \textbf{\color{finsta-color} Finsta}-Clover & 97.9\textcolor{mred}{\scriptsize{\textbf{ +2.9}}}& 47.8\textcolor{mred}{\scriptsize{\textbf{ +5.3}}}& 54.6\textcolor{mred}{\scriptsize{\textbf{ +3.5}}}&  75.8\textcolor{mred}{\scriptsize{\textbf{ +4.2}}} \\

All-in-one & 92.3 & 44.3 & 47.9  & 66.0 \\
\rowcolor{blue} \textbf{\color{finsta-color} Finsta}-All-in-one & 94.0\textcolor{mred}{\scriptsize{\textbf{ +1.7}}}& 45.2\textcolor{mred}{\scriptsize{\textbf{ +0.9}}}& 50.0\textcolor{mred}{\scriptsize{\textbf{ +2.1}}}& 68.1\textcolor{mred}{\scriptsize{\textbf{ +2.1}}}\\

\hdashline
Video-LLaMA & 95.3 & 48.3 & 57.2  & 75.3  \\
\rowcolor{blue} \textbf{\color{finsta-color} Finsta}-Video-LLaMA & 98.1\textcolor{mred}{\scriptsize{\textbf{ +2.8}}}& \color{mblue} \bf 51.3\textcolor{mred}{\scriptsize{\textbf{ +3.0}}}& 62.7\textcolor{mred}{\scriptsize{\textbf{ +5.5}}}& 82.4\textcolor{mred}{\scriptsize{\textbf{ +7.1}}}\\

Video-LLaVA & 96.8 & 48.0 & \underline{65.7}  & 77.5  \\
\rowcolor{blue} \textbf{\color{finsta-color} Finsta}-Video-LLaVA & \color{mblue} \bf 99.4\textcolor{mred}{\scriptsize{\textbf{ +2.6}}}& 51.7\textcolor{mred}{\scriptsize{\textbf{ +3.7}}}& \color{mblue} \bf 72.5\textcolor{mred}{\scriptsize{\textbf{ +6.8}}}& \color{mblue} \bf 83.1\textcolor{mred}{\scriptsize{\textbf{ +5.6}}}\\

\hline
\end{tabular}
\vspace{-3mm}
\label{tab:VQA}
\end{table}

\vspace{-2mm}
\subsection{Main Results on VL Modeling Tasks}
\label{Experimental Results and Discussions}

We first evaluate the fine-tuning performance of Finsta on a wide range of VL tasks.\footnote{
Due to the space limit, we present the complete results on more datasets with more metrics for each task in Supplementary Section 4.
}

\vspace{-2mm}
\subsubsection{Results on Video-to-Text Transformation Tasks}

\noindent\textbf{$\blacktriangleright$ Video Action Recognition.}
Table \ref{tab:VAR} presents the overall performance of the K400 and SSV2 datasets, where the Finsta-enhanced VLMs also compare with the existing strong-performing systems.
As seen, InternVideo has been the existing state-of-the-art system for the task.
However, our Finsta further helps InternVideo improve by 2.6\% and 3.3\% Top-1 accuracy on two datasets, respectively, which makes the Finsta-InternVideo the current new state-of-the-art on both two benchmarks.
Overall, all the VLMs witness performance increases to different extents.
Notably, HDVILA is enhanced by an average of above 4\% accuracy.

\vspace{1mm}
\noindent\textbf{$\blacktriangleright$ Video Captioning.}
Table \ref{tab:VC} shows the results on three benchmarks.
First of all, we can see that all different VLMs receive varied improvements from Finsta with clear margins.
With Finsta, both the state-of-the-art performances on two datasets further have been refreshed.
This validates the effectiveness of our method for enhancing video-to-language type of task understanding.
Also, besides the dual-stream-sum VLMs, the All-in-one VLM which only has one shared multimodal encoder is included.
Interestingly, compared with other combinations, Finsta-All-in-one receives the least improvements.
This is largely because of the absence of the two key modules: the textual encoder and video encoder.

\vspace{-2mm}
\subsubsection{Results on Text-to-Video Transformation Task}

\noindent\textbf{$\blacktriangleright$ Video-Text Retrieval.}
We use a total of 9 VLMs as backbone, where we further compare with 5 strong-performing baselines.
We present the overall results in Table \ref{tab:VTR}.
As seen, our Finsta still consistently improves all the VLMs by clear margins.
Among them, the Finsta-Video-LLaMA and the Finsta-Video-LLaVA have been the new state-of-the-arts on all the datasets.
Notably, we find that HDVILA benefits from Finsta the most, with an average of 10.3\% recall rate.
Likewise, the multimodal-encoder-only All-in-one VLM has received the least boosts from Finsta.
Also, without the cross-modal encoder, the enhancements for VideoCLIP and CLIP4Clip VLMs are limited, compared with all the rest VLMs with a full dual-stream-sum architecture.

\vspace{-2mm}
\subsubsection{Results on Video-Text Collaboration Task}

\noindent\textbf{$\blacktriangleright$ Video Question Answering.}
Table \ref{tab:VQA} presents the results on both the multi-choice QA and open-ended QA.
Similar to the aforementioned tasks, all different backbone VLMs are improved via our Finsta system.
Among them, the Finsta-Video-LLaMA and the Finsta-Video-LLaVA surpass all the existing performances, setting new state-of-the-art records on all the datasets in different QA settings.
Most significantly, Finsta boosts Video-LLaVA by 6.8\% accuracy in MSVD-QA data.
Also there is a similar trend for the All-in-one VLM, which gets the most conservative improvement.

\begin{table}[t]
\centering
\caption{Video-Paragraph Retrieval results on two datasets.}
\vspace{-2mm}
\fontsize{9}{10}\selectfont
\setlength{\tabcolsep}{1.2mm}
\begin{tabular}{lllll}
\hline
\multicolumn{1}{c}{\multirow{2}{*}{\bf Method}} & \multicolumn{2}{c}{\bf QuerYD \cite{OncescuHLZA21}} & \multicolumn{2}{c}{\bf ActivityNet \cite{KrishnaHRFN17}}\\
\cmidrule(r){2-3}\cmidrule(r){4-5}
 & \bf R@1 & \bf R@5 & \bf R@1 & \bf R@5 \\
\midrule

TeachText \cite{CroitoruBLJZAL21} & 14.4 & 37.7 & - & - \\
Frozen \cite{BainNVZ21} & 53.8 & 75.7 & 28.8 & 60.9 \\
TESTA \cite{ren2023testa}& \underline{83.4} & \underline{93.8} & {54.8} & {80.8} \\

\hline
LFVILA & 69.7 & 85.7 & 35.3 & 65.4 \\
\rowcolor{blue} \textbf{\color{finsta-color} Finsta}-LFVILA (S-Vid) & 72.0\textcolor{mred}{\scriptsize{\textbf{ +2.3}}} & 87.4\textcolor{mred}{\scriptsize{\textbf{ +1.7}}} & 37.7\textcolor{mred}{\scriptsize{\textbf{ +2.4}}} & 68.0\textcolor{mred}{\scriptsize{\textbf{ +2.6}}} \\
\rowcolor{blue} \textbf{\color{finsta-color} Finsta}-LFVILA (L-Vid) & 78.6\textcolor{mred}{\scriptsize{\textbf{ +8.9}}} & \color{mblue} \bf 94.5\textcolor{mred}{\scriptsize{\textbf{ +8.8}}} & 69.8\textcolor{mred}{\scriptsize{\textbf{ +34.5}}} & \color{mblue} \bf 92.5\textcolor{mred}{\scriptsize{\textbf{ +27.1}}} \\

\hdashline
Video-LLaMA & 71.5 & 86.2 & 49.4 & 68.4 \\
\rowcolor{blue} \textbf{\color{finsta-color} Finsta}-Video-LLaMA & 78.5\textcolor{mred}{\scriptsize{\textbf{ +7.0}}} & 90.2\textcolor{mred}{\scriptsize{\textbf{ +4.0}}} & 56.5\textcolor{mred}{\scriptsize{\textbf{ +7.1}}} & 73.8\textcolor{mred}{\scriptsize{\textbf{ +5.4}}} \\

Video-ChatGPT & 84.4 & 90.5 & 66.8 & 86.1 \\
\rowcolor{blue} \textbf{\color{finsta-color} Finsta}-Video-ChatGPT & \color{mblue} \bf 86.8\textcolor{mred}{\scriptsize{\textbf{ +2.4}}} & 94.1\textcolor{mred}{\scriptsize{\textbf{ +3.6}}} & 69.3\textcolor{mred}{\scriptsize{\textbf{ +2.5}}} & 89.7\textcolor{mred}{\scriptsize{\textbf{ +3.6}}} \\

Video-LLaVA & 76.8 & 88.0 & \underline{68.4} & \underline{89.3} \\
\rowcolor{blue} \textbf{\color{finsta-color} Finsta}-Video-LLaVA & 80.1\textcolor{mred}{\scriptsize{\textbf{ +3.3}}} & 91.0\textcolor{mred}{\scriptsize{\textbf{ +3.0}}} & \color{mblue} \bf 71.6\textcolor{mred}{\scriptsize{\textbf{ +3.2}}} & 91.8\textcolor{mred}{\scriptsize{\textbf{ +2.5}}} \\

\hline
\end{tabular}
\label{tab:VPR}
\end{table}

\begin{table}[t]
\centering
\caption{Long-Form Video Question-Answering results (Acc.) on two datasets.
}
\fontsize{9}{10}\selectfont
\setlength{\tabcolsep}{2mm}
\begin{tabular}{lll}
\hline
\bf Method & \bf How2QA \cite{li-etal-2020-hero}	& \bf VIOLIN \cite{LiuCCGYYL20}\\   
\midrule
ResNet-SF \cite{LiLGY0P0ZWWBB0W21} & 	74.3 & 	- \\   
GVE \cite{ChenG21} & 	-	 & 68.4 \\   
HERO \cite{li-etal-2020-hero} & 	74.3 & 	68.6 \\   
\cline{1-3}
LFVILA & 	76.1 & 	70.9 \\
\rowcolor{blue} \textbf{\color{finsta-color} Finsta}-LFVILA (S-Vid) & 77.5\textcolor{mred}{\scriptsize{\textbf{ +1.4}}} & 71.7\textcolor{mred}{\scriptsize{\textbf{ +0.8}}} \\
\rowcolor{blue} \textbf{\color{finsta-color} Finsta}-LFVILA (L-Vid) & 84.8\textcolor{mred}{\scriptsize{\textbf{ +8.7}}} & \color{mblue} 	\bf 78.0\textcolor{mred}{\scriptsize{\textbf{ +7.1}}} \\

\hdashline
Video-LLaMA   & 79.6 & \underline{75.3} \\
\rowcolor{blue} \textbf{\color{finsta-color} Finsta}-Video-LLaMA  & 84.0\textcolor{mred}{\scriptsize{\textbf{ +4.4}}} & 77.8\textcolor{mred}{\scriptsize{\textbf{ +2.5}}} \\

Video-ChatGPT   & 80.7 & 73.4 \\
\rowcolor{blue} \textbf{\color{finsta-color} Finsta}-Video-ChatGPT  & 84.5\textcolor{mred}{\scriptsize{\textbf{ +3.8}}} & 76.9\textcolor{mred}{\scriptsize{\textbf{ +3.5}}} \\

Video-LLaVA   & \underline{83.5} & 75.0 \\
\rowcolor{blue} \textbf{\color{finsta-color} Finsta}-Video-LLaVA  & \color{mblue} 	\bf  87.8\textcolor{mred}{\scriptsize{\textbf{ +4.3}}} & 77.3\textcolor{mred}{\scriptsize{\textbf{ +2.3}}} \\

\hline
\end{tabular}
\vspace{-3mm}
\label{tab:LFVQA}
\end{table}

\vspace{-2mm}
\subsubsection{Results on Long-form Video-Text Tasks}

\noindent\textbf{$\blacktriangleright$ Video-Paragraph Retrieval.}
As presented in Table \ref{tab:VPR}, Finsta helps all different VLMs achieve different levels of improvement, in which two points can be observed.
First, we see that the LFVILA (L-Vid) has shown much stronger performance than the LFVILA (S-Vid).
Also, Finsta-LFVILA (L-Vid) becomes the new state-of-the-art on two datasets.
Such a gap indicates the importance of training VLMs with long-form videos (and texts) for long-form scenarios.
Besides, among three different LVLMs, Video-LLaMA receives the most significant boost from Finsta.

\vspace{2mm}
\noindent\textbf{$\blacktriangleright$ Long-form Video Question Answering.}
Table \ref{tab:LFVQA} reports the results on two datasets.
Likewise, all four different systems are enhanced consistently, with the L-Vid LFVILA obtaining the biggest improvements.
Finsta strikingly boosts the raw LFVILA (L-Vid) VLM by 8.7\% and 7.1\% accuracy on two datasets, respectively.
The Finsta-Video-LLaVA and Finsta-LFVILA (L-Vid) become the new state-of-the-arts on two datasets, respectively.
The above consistent improvement of our Finsta system on the long-form setting evidently verifies its efficacy in improving the VL modeling and understanding.


\begin{table}[t]
\centering
\caption{Zero-shot results of \colorbox{nmdeepred}{Video Action Recognition}, \colorbox{nmdeepyellow}{Video-Text Retrieval} and \colorbox{nmdeepgreen}{Video Question Answering}.}
\vspace{-2mm}
\fontsize{9}{10}\selectfont
\setlength{\tabcolsep}{0.5mm}
\begin{tabular}{lllll}
\hline
\multicolumn{1}{c}{\multirow{2}{*}{\bf Method}}& \multicolumn{1}{>{\columncolor{nmdeepred}}c}{\bf K400 \cite{KayCSZHVVGBNSZ17}} & \multicolumn{2}{>{\columncolor{nmdeepyellow}}c}{\bf LSMDC \cite{MaharajBRCP17}} &   \multicolumn{1}{>{\columncolor{nmdeepgreen}}c}{\bf MSVD \cite{chen-dolan-2011-collecting}}\\
 & \multicolumn{1}{>{\columncolor{nmred}}c}{\bf Top-1} & \multicolumn{1}{>{\columncolor{nmyellow}}c}{\bf R@1} & \multicolumn{1}{>{\columncolor{nmyellow}}c}{\bf R@5} &\multicolumn{1}{>{\columncolor{nmgreen}}c}{\bf OE} \\
\midrule

Frozen \cite{BainNVZ21} & 	-  & 9.3 & 22.0  &  8.9 \\
OmniVL \cite{Wang0WLZZX0JY22} & -& 8.6 & 18.7&  - \\
HiTeA \cite{ye2023hitea} & - &15.5& \underline{31.1} &  {18.2} \\
ImageBind \cite{girdhar2023imagebind}& 50.0 & 10.4 & 27.9 &  14.5 \\
LanguageBind \cite{zhu2023languagebind} & {64.0} & \underline{16.4} & 30.8 &  12.3 \\

\hline
VIOLET & 44.8 & 8.6 & 23.2 & 11.0 \\
\rowcolor{blue} \textbf{\color{finsta-color} Finsta}-VIOLET & 59.7\textcolor{mred}{\scriptsize{\textbf{ +14.9}}}& 15.5\textcolor{mred}{\scriptsize{\textbf{ +6.9}}}& 35.5\textcolor{mred}{\scriptsize{\textbf{ +12.3}}}&  27.2\textcolor{mred}{\scriptsize{\textbf{ +16.2}}} \\

VideoCLIP & 36.4 & 9.5 & 16.3  & 7.6 \\
\rowcolor{blue} \textbf{\color{finsta-color} Finsta}-VideoCLIP & 46.7\textcolor{mred}{\scriptsize{\textbf{ +10.3}}}& 12.4\textcolor{mred}{\scriptsize{\textbf{ +2.9}}}& 19.8\textcolor{mred}{\scriptsize{\textbf{ +3.5}}}& 15.7\textcolor{mred}{\scriptsize{\textbf{ +8.1}}} \\

CLIP4Clip & 48.3 & 11.3 & 22.7   & 10.6 \\
\rowcolor{blue} \textbf{\color{finsta-color} Finsta}-CLIP4Clip & 54.5\textcolor{mred}{\scriptsize{\textbf{ +6.2}}}& 13.9\textcolor{mred}{\scriptsize{\textbf{ +2.6}}}& 26.7\textcolor{mred}{\scriptsize{\textbf{ +4.0}}}& 18.5\textcolor{mred}{\scriptsize{\textbf{ +7.9}}} \\

MCQ & 41.3 & 12.2  & 25.9 &  8.2 \\
\rowcolor{blue} \textbf{\color{finsta-color} Finsta}-MCQ & 55.2\textcolor{mred}{\scriptsize{\textbf{ +13.9}}} & 15.8\textcolor{mred}{\scriptsize{\textbf{ +3.6}}} & 37.5\textcolor{mred}{\scriptsize{\textbf{ +11.6}}} & 25.8\textcolor{mred}{\scriptsize{\textbf{ +17.6}}} \\

InternVideo & 53.3 & 11.0 & 25.8 & 13.5 \\
\rowcolor{blue} \textbf{\color{finsta-color} Finsta}-InternVideo & 68.9\textcolor{mred}{\scriptsize{\textbf{ +15.6}}} & 21.9\textcolor{mred}{\scriptsize{\textbf{ +10.9}}} & 39.3\textcolor{mred}{\scriptsize{\textbf{ +13.5}}} &  29.8\textcolor{mred}{\scriptsize{\textbf{ +16.3}}} \\

\hdashline
Video-LLaMA & 60.3 & 13.5 & 26.3  & 20.3 \\
\rowcolor{blue} \textbf{\color{finsta-color} Finsta}-Video-LLaMA & 69.7\textcolor{mred}{\scriptsize{\textbf{ +9.4}}} &\color{mblue} \bf 23.9\textcolor{mred}{\scriptsize{\textbf{ +10.4}}} & 37.8\textcolor{mred}{\scriptsize{\textbf{ +11.5}}} & 36.6\textcolor{mred}{\scriptsize{\textbf{ +16.3}}} \\

Video-LLaVA & \underline{64.3} & 15.5 & 29.8 &  \underline{25.1} \\
\rowcolor{blue} \textbf{\color{finsta-color} Finsta}-Video-LLaVA & \color{mblue} \bf 72.2\textcolor{mred}{\scriptsize{\textbf{ +7.9}}} & 22.0\textcolor{mred}{\scriptsize{\textbf{ +6.5}}} & \color{mblue} \bf 43.5\textcolor{mred}{\scriptsize{\textbf{ +13.7}}} & \color{mblue} \bf 45.3\textcolor{mred}{\scriptsize{\textbf{ +20.2}}} \\

\hline
\end{tabular}
\vspace{-3mm}
\label{zero-shot}
\end{table}


\vspace{-2mm}
\subsection{Zero-shot Video-Language Understanding Results}

Here we examine Finsta's efficacy in zero-shot setting, where VLMs make predictions on the downstream VL tasks without fine-tuning the on-demand training data.
We representatively test on three VL tasks: Video Action Recognition, Video-Text Retrieval and Video Question Answering, each using different dataset(s).
The results are presented in Table \ref{zero-shot}, from which we can gain several key observations.
First of all, all VLMs have shown significant improvement by Finsta, notably with Finsta-InternVideo and Finsta-Video-LLaVA emerging as the new best in zero-shot state-of-the-art performance. 
Second, when compared to the prior fine-tuning results, the enhancements brought by Finsta to zero-shot scenarios are strikingly more evident. 
This underscores Finsta's role in fine-grained structured VL alignment learning, essentially providing a crucial signal that aids unsupervised learning.
Finsta takes full advantage of the external semantic scene structural features (i.e., SGs) for enhancement of VL understanding, which correspondingly gives rise to these improvements.
Lastly, in contrast to other VLMs with full dual-stream-sum architecture, the lowest improvements of VideoCLIP and CLIP4Clip suggest that the absence of certain modules can significantly impact the performance of Finsta.

\vspace{-2mm}
\section{Discussions and In-depth Analyses}

In this section, we give more discussions via a series of in-depth analyses to reveal how the system advances.\footnote{
Due to the space limit, we present more experiments and analysis results
in Supplementary Material Section 5.
}
Following, we try to ground the answers for the following five key research questions:

\vspace{1mm}
\textbf{RQ-1: Does Finsta improve VLMs by really addressing the bottlenecks of VLMs?}

\vspace{1mm}
\textbf{RQ-2: How much does each module contribute to the overall Finsta?}

\vspace{1mm}
\textbf{RQ-3: What factors influence Finsta's performance? How do different factors impact Finsta?}

\vspace{1mm}
\textbf{RQ-4: What is the Finsta's computational cost? Is Finsta efficient compared to its efficacy?}

\vspace{1mm}
\textbf{RQ-5: How exactly does Finsta advance better VLMs?}

\vspace{-2mm}
\subsection{RQ-1: Does Finsta Address VLM Bottlenecks?}

\begin{figure}[t]
\centering
\includegraphics[width=0.98\columnwidth]{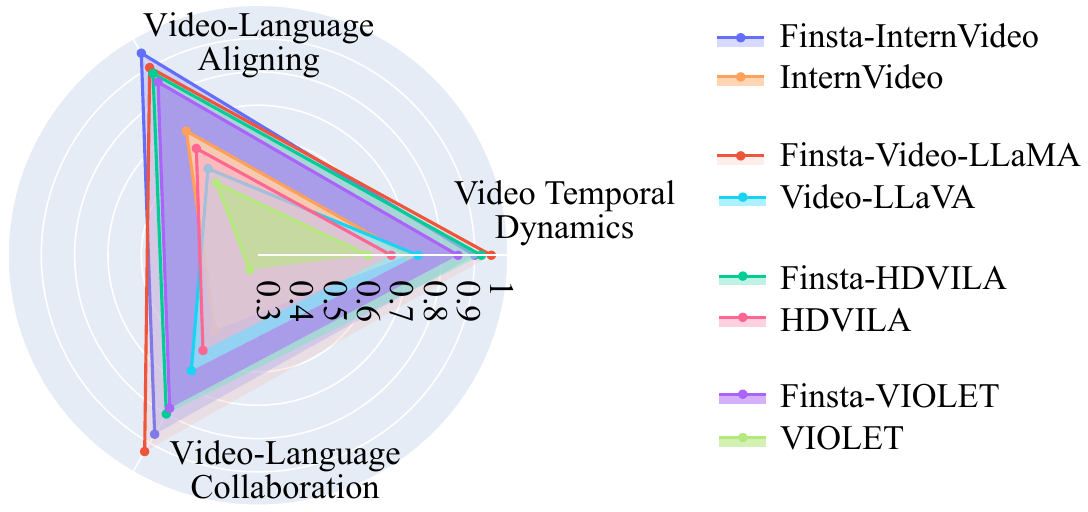}
\caption{
Video-Paragraph Retrieval results with varying numbers of video actions in ActivityNet data.
}
\label{fig:hostic-assess}
\end{figure}

In the Introduction section we highlight the imperatives of resolving three key bottlenecks of existing strong-performing VLMs in modeling video\&language.
Here we investigate if the enhancement of Finsta over backbone VLMs is really coming from addressing these main challenges.
To directly answer the question, we consider a human evaluation of the VL modeling.
We select four VLMs, including VIOLET, HDVILA, InternVideo and Video-LLaVA.
We sampled 300 long video-language pairs from ActivityNet \cite{KrishnaHRFN17} test set for zero-shot Video Captioning, Question Answering and VL Retrieval tasks, with each video containing more than 8 actions for imitating the challenging action-complex video scenario.
We ask 5 trained volunteers to assess the direct performance of each VLMs in terms of \emph{Video-Language Aligning}, \emph{Video Temporal Dynamics} and \emph{Video-Language Collaboration} based on the VLMs' outputs.
In Figure \ref{fig:hostic-assess} we plot the change before and after equipping with Finsta, where Finsta evidently enhances the abilities in these three aspects.


\begin{figure}[!t]
\begin{center}
\includegraphics[width=.96\linewidth]{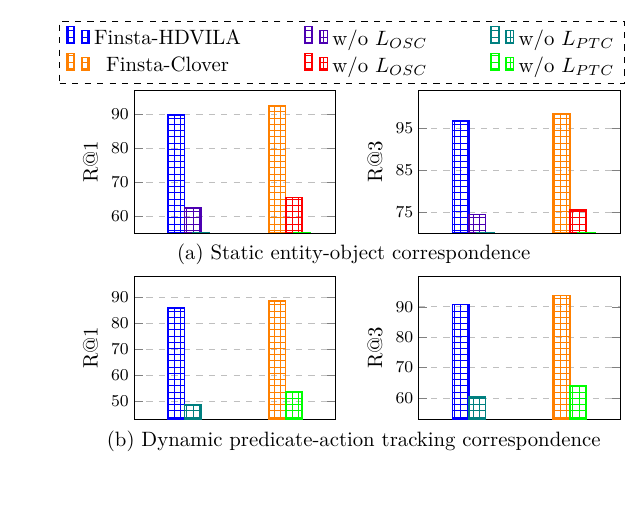}
\end{center}
\vspace{-2mm}
\caption{
Evaluation of fine-grained TSG and DSG node correspondence.
}
\label{app-correspd}
\vspace{-3mm}
\end{figure}

Further, we step further in evaluating how better the fine-grained video-language alignment/grounding Finsta has been achieved.
We probe the exact correspondences between the two modalities, in terms of the static token-object alignment, and the dynamic predicate-action tracking alignment, respectively.
We measure each pair of nodes from the TSG and DSG respectively, i.e., the bipartite scores: $S^o_{i,t,j}( v^T_i, v^D_{t,j} )$, and $S^p_{i,t:t+m,j} ( v^T_i,  v^D_{t:t+m,j} )$.
Note that here we degrade the order $n$ as 0, i.e., only considering the object node itself.
We mainly examine the HDVILA and Clover VLMs equipped with our Finsta.
The results are shown in Figure \ref{app-correspd}.
As seen, both 1) the spatial textual-entity and visual-object alignments and 2) dynamic temporal predicate-action grounding are well captured in Finsta system.


\begin{figure}[!t]
\centering
\includegraphics[width=0.98\columnwidth]{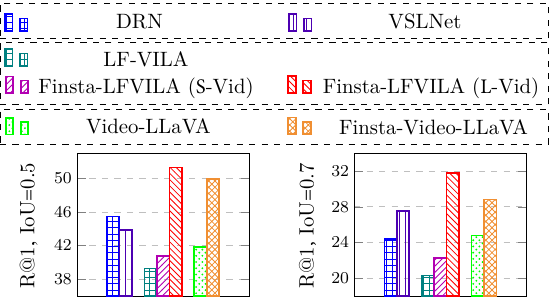}
\caption{
Language Video Localization on ActivityNet data.
}
\vspace{-3mm}
\label{fig:vid-loc}
\end{figure}

One direct assessment of the temporal dynamics modeling is through the Language Video Localization task, which requires localizing the specific temporal moments in an untrimmed video precisely by a given language query.
We experiment with the same challenging ActivityNet data, and compare our VLMs with the state-of-the-art models including DRN \cite{zeng2020dense} and VSLNet \cite{zhang2021natural}.
As shown in Figure \ref{fig:vid-loc}, our Finsta boosts the backbone VLMs with striking gains in grounding the video temporality from language.

\begin{table*}[t]
\centering
\caption{Model ablation of Finsta-HDVILA and Finsta-LFVILA on different datasets.
}
\fontsize{9}{10}\selectfont
\setlength{\tabcolsep}{1.5mm}
\begin{tabular}{lccccccl}
\hline
& \multicolumn{4}{c}{ \bf  HDVILA} & \multicolumn{2}{c}{ \bf  LFVILA (L-Vid)} & \multicolumn{1}{c}{\multirow{3}{*}{\bf AVG}}\\
\cmidrule(r){2-5}\cmidrule(r){6-7}
 & \bf \specialcell{K400\\{\scriptsize(VAR, Top-1)}} & \bf \specialcell{MSR-VTT\\{\scriptsize(VC, M)}} & \bf \specialcell{DiDeMo\\{\scriptsize{(VTR. R@1)}}}	 & \bf \specialcell{MSVD\\{\scriptsize(VQA, Acc.)}} 
& \bf \specialcell{ActivityNet\\{\scriptsize(VPR, R@1)}}
& \bf \specialcell{How2QA\\{\scriptsize(LF-VQA, Acc.)}} \\   
\midrule
Finsta-[\emph{VLM}] & \bf 83.4& \bf 36.9 & \bf 49.3 & \bf 53.3 &\bf 69.8 &\bf 84.8 & \bf 62.9 \\

\cdashline{1-8}
\multicolumn{4}{l}{\textbf{$\blacktriangleright$ SG Representation Integration}} \\
\quad w/o Temporal coref. edge in DSG & 82.1&	33.3 & 	47.0 & 	52.2 & 59.7 & 81.0& 59.2\scriptsize{\textbf{ -3.7 }}\\  
\quad w/o Adverbial modifier in TSG & 83.0&	34.0 &	47.2 &	51.9 & 61.1& 82.9& 60.0\scriptsize{\textbf{ -2.9 }}\\   
\quad w/o X-modal coref. edge in HSG & 82.7&	35.0 & 	48.2 & 	52.6 & 62.4 & 82.0 & 60.5\scriptsize{\textbf{ -2.4 }}\\   
\cdashline{1-8}
\multicolumn{4}{l}{\textbf{$\blacktriangleright$ SG Encoding}} \\
\quad GTrm$\to$GAT & 82.5 &36.2 &	48.9 &	53.0 & 57.7 & 81.5& 59.9\scriptsize{\textbf{ -2.9 }}\\  
\quad R-GTrm$\to$RGNN  & 82.0 &35.2 &	48.5 &	52.7 & 55.3 & 80.1& 58.9\scriptsize{\textbf{ -3.9 }}\\  
\qquad w/o STGD-GTrm &	81.7 & 35.4 & 47.2 &	52.0  & 52.1& 78.6& 57.8\scriptsize{\textbf{ -5.1 }}\\  

\cdashline{1-8}
\multicolumn{4}{l}{\textbf{$\blacktriangleright$ Alignment Learning}} \\
\quad w/o $L_{\text{OSC}}$ & 80.9 &33.4 &	46.8 &	51.6 & 50.9 & 78.6& 57.0\scriptsize{\textbf{ -5.9 }}\\ 
\qquad w/o high-order &	83.0 & 35.9 & 48.8 &	53.1 & 57.0& 79.7 & 59.6\scriptsize{\textbf{ -3.3 }}\\  
\quad w/o $L_{\text{PTC}}$ & 79.5 &32.8 & 	46.1 & 	51.0 & 43.5 & 77.2& 55.0\scriptsize{\textbf{ -7.9 }}\\  
\qquad w/o high-order &	82.6 & 34.0 & 47.3 &	51.9 & 52.4& 78.9& 57.8\scriptsize{\textbf{ -5.1 }}\\   
\hline
\end{tabular}
\vspace{-3mm}
\label{tab:Ablation}
\end{table*}

\vspace{-2mm}
\subsection{RQ-2: How Much Does Each Module Contribute?}

To understand the exact contribution of each component, here we present an ablation study.
We analyze Finsta-HDVILA and Finsta-LFVILA on three aspects: SG representations, SG encoders and alignment learning.
The results are shown in Table \ref{tab:Ablation}.
First, by 1) canceling the adverbial modifier node in TSG,
2) removing the temporal coreference edges in DSG 
and 3) dropping the cross-modal coreference edges in HSG, respectively,
we can witness different levels of performance decreases.
Among them, the installation of temporal coreference edges for DSG is of the most importance, i.e., average -3.7\%.
Further, we replace the Transformer-based GTrm with the GAT \cite{VelickovicCCRLB18} for TSG encoding, and replace the R-GTrm with RGNN \cite{NicolicioiuDL19} encoder for the DSG \& HSG encoding, respectively.
There are also considerable performance drops accordingly, indicating the effectiveness of using the GTrm and R-GTrm encoders.
More significantly, we see that by removing the STGD-GTrm, we witness a drop of -5.1\%, highlighting the importance of modeling the moving changes of videos.
Finally, we cancel the alignment learning of either the object-centered spatial contrasting ($L_{\text{OSC}}$) or the predicate-centered temporal contrasting ($L_{\text{PTC}}$), there is the most significant performance downgrading, compared with any other factors.
Especially the temporal alignment ($L_{\text{PTC}}$) shows the most striking influences among all the rest modules.
Further, if we downgrade the high-order feature modeling of VL alignment into the first-order manner, we can also see considerable drops.
This certifies the prominence of the high-order feature modeling.

\begin{figure}[!t]
\begin{center}
\includegraphics[width=1\columnwidth]{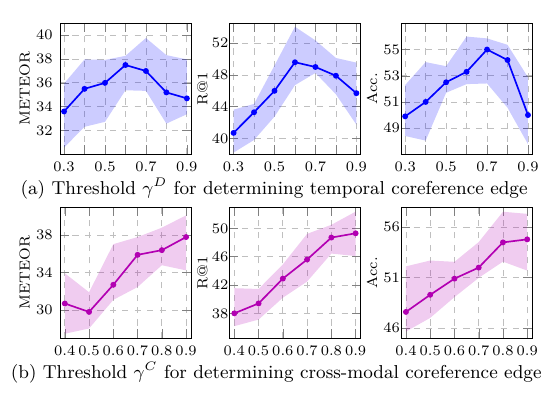}
\end{center}
\vspace{-3mm}
\caption{
Analysis of the impact of threshold $\gamma^D$ and $\gamma^C$.
}
\vspace{-3mm}
\label{gamma-analysis}
\end{figure}

\vspace{-2mm}
\subsection{RQ-3: What Factors Influence Finsta?}

In this part, we analyze all possible factors that will influence Finsta's performance.

\vspace{-2mm}
\subsubsection{Influence of Hyperparameters}

We mainly study three key sets of hyperparameters: the $\gamma^D$ \& $\gamma^C$ in constructing temporal and cross-modal coreference edges in DSG and HSG;
the orders of neighbor features for OSC and PTC alignment learning;
and the alignment confidence thresholds $\rho^o$ \& $\rho^p$.

\vspace{1mm}
\noindent\textbf{1) Influence of Different Thresholds for Building Temporal and Cross-Modal Coreference Edge.}
We vary the values of $\gamma^D$ and $\gamma^C$, and use the data of constructed SGs for training the VLMs, and then explore the performance of Finsta-HDVILA on various VL tasks.
In Figure \ref{gamma-analysis}, we plot the results.
As seen, different values of $\gamma^D$ and $\gamma^C$ lead to distinct results of end tasks.
We find that when $\gamma^D$ is set as 0.6, the best quality of temporal coreference edges in DSG can be obtained.
And setting $\gamma^C$ to 0.9 seems to be optimal for building the cross-modal coreference edges in HSG.
This is because the video and language modalities come with bigger gaps in semantics, which naturally demands a larger threshold for finding their match.

\begin{figure}[!t]
\begin{center}
\includegraphics[width=0.99\columnwidth]{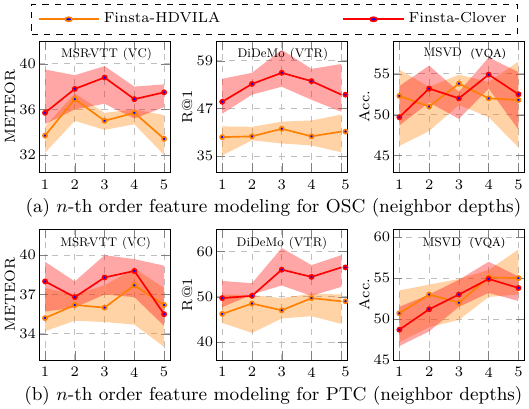}
\end{center}
\vspace{-3mm}
\caption{
Finsta-HDVILA and Finsta-Clover performance with different $n$-order neighboring contexts.
}
\label{high-order}
\end{figure}

\begin{figure}[!t]
\begin{center}
\includegraphics[width=0.99\columnwidth]{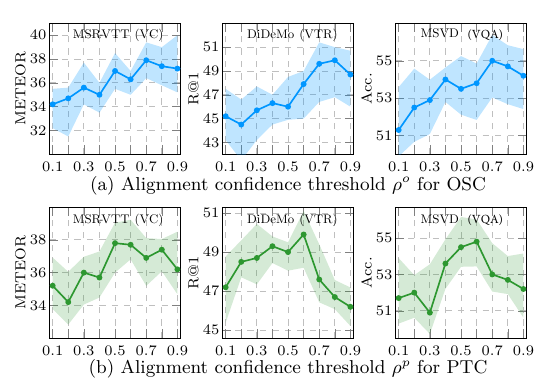}
\end{center}
\vspace{-3mm}
\caption{
Finsta-HDVILA performance on different tasks with varied alignment threshold $\rho^o$ and $\rho^p$.
}
\vspace{-3mm}
\label{threshold}
\end{figure}

\begin{figure}[t]
\begin{center}
\includegraphics[width=0.99\columnwidth]{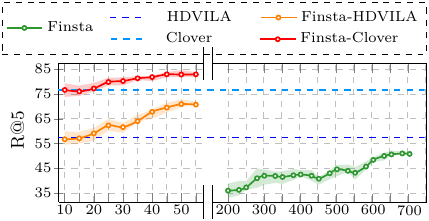}
\vspace{-4mm}
\end{center}
\caption{
Finsta-VLM performance on the Video-Text Retrieval task (DiDeMo) by post-training with varied SG data amount.
}
\label{data-amount}
\end{figure}

\vspace{1mm}
\noindent\textbf{2) Influence of High-order Neighboring Modeling.}
Intuitively, higher-order ($n$) of feature modeling allows larger context windows, yet at the cost of covering more noises.
In Figure \ref{high-order} we present the trends over both Finsta-HDVILA and Finsta-Clover models.
As seen, mostly the OSC learning relies on a 3rd-order feature for the static spatial alignment;
while for the PTC learning, a 4th-order context of regions is needed for the best dynamic temporal alignment.
This is reasonable as the alignment learning for temporal dynamics modeling depends more on the contexts of two modalities.

\vspace{1mm}
\noindent\textbf{3) Influence of Different Threshold Value for Alignment Learning.}
We further probe the impact of setting different thresholds $\rho^o$ and $\rho^p$ for the fine-grained spatial and temporal alignment learning. 
In Figure \ref{threshold} we present the results.
We see that the trends of $\rho^o$ and $\rho^p$ can be slightly different.
Overall, The best values of $\rho^o$ are higher than that of $\rho^p$.
This indicates that, the alignments of temporal dynamics require more shreds of evidence between the two modalities.
In summary, we set $\rho^o$ as 0.7 for OSC, and $\rho^p$ as 0.6 for PTC, where we can secure the best performance.

\vspace{-2mm}
\subsubsection{Influence of Post-training Data Amount}
One general viewpoint is that, the larger the data used for training, the better the resulting performance.
For post-training Finsta, we used quite a limited amount of data, i.e., a total of 50K for normal-scene videos, which is only 0.94\% for pre-training the raw Clover (with 5.3M), and 0.037\% for pre-training the raw HDVILA (with 136M).
In Figure \ref{data-amount} we verify this claim by evaluating the end task (Video-Text Retrieval) performance with Finsta-VLM post-trianed using different numbers of SG data.
As seen, even using SG data not reaching 50K, both the Finsta-HDVILA and Finsta-Clover can quickly climb to their best performances, based on the pre-trained HDVILA and Clover VLMs.
This is because the well pre-trained backbone VLMs provide a warm starting for more rapid convergence of training.
However, if we treat the Finsta as a standalone VLM, and (pre-)train it from scratch with SG annotations, we find that the pre-training process requires much more data, and also results in lower peak.

\begin{figure}[!t]
\begin{center}
\includegraphics[width=1\linewidth]{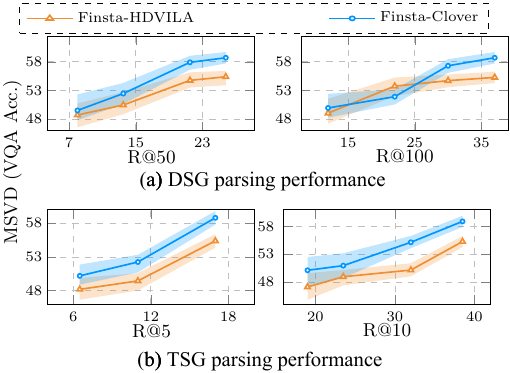}
\end{center}
\vspace{-3mm}
\caption{
Influences of the SG parser quality.
}
\vspace{-3mm}
\label{SG-quality}
\end{figure}

\begin{figure}[!t]
\begin{center}
\includegraphics[width=1\linewidth]{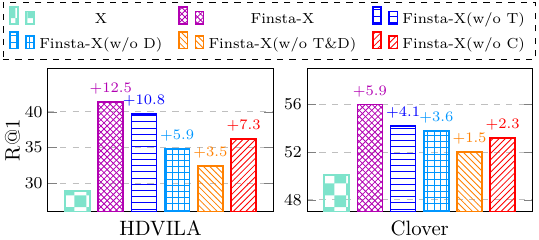}
\end{center}
\vspace{-2mm}
\caption{
Influence of absence of certain VLM modules.
}
\vspace{-3mm}
\label{module-absence}
\end{figure}

\vspace{-2mm}
\subsubsection{Influence of SG Parsing Quality}

Our proposed Finsta system relies much on the availability of the SGs.
Here we study how the SG parser quality affects the final results.
We consider varying the performance of the scene graph generation (SGG) step in SG parsing.
We vary these parser qualities by training them with early existing, such that they will show different developing performances.
The conventional metric for the evaluation of SGG is Recall@K (R@K).
In Figure \ref{SG-quality} we show the end task results (VQA on MSVD) of Finsta-VLMs that are post-trained with different qualities of SG data.
Overall, the lower quality of SG annotations does hurt the VL learning in the resulting Finsta.
Also, we find that the Finsta performances are more sensitive to the quality of TSG.
Intuitively, lower-quality SG structures provide false supervision that will mislead the fine-grained structural alignment.
Fortunately, the quality of the SG parser currently in use is sufficient to obtain satisfactory SG annotations, as evidenced in our current implementation. 
Most importantly, our Finsta system relies only on a minimal amount of SG annotated data for post-training.
During the end-task fine-tuning, there is actually no need to feed in additional SGs, effectively avoiding the noise introduced by SG parsing quality issues in the end-task phase.

\vspace{-2mm}
\subsubsection{Influence of VLM Module Existence}
\label{Influence of VLM Module Existence}

The plug\&play design of Finsta can be convenient to be applied to any other existing VLMs.
While Finsta has a dual-stream-sum architecture (i.e., text/video/cross-modal encoders), 
even if any of the three encoders are absent, Finsta can still work.
However, without a complete architecture, the efficacy of Finsta will be intrinsically sacrificed to a certain extent.
This can be directly observed in the above experiments of VL end tasks, on  VideoCLIP, CLIP4Clip and All-in-one.
To give a direct sense, here we study a VLM that originally has full dual-stream-sum structures, and we ablate it by removing certain modules and then equipping it with Finsta system.
In Figure \ref{module-absence} we see that removing any part of the Finsta modules from injecting into VLMs can lead to clear drops.
Especially canceling 1) simultaneously the textual\&video encoder (TSG\&DSG encoder) for VL fine-grained alignment, or 2) the multimodal encoder (HSD modeling), the decreases are much more evident.
Fortunately, most of the existing VLMs come with a dual-stream-sum architecture, where our Finsta can work sufficiently.

\begin{table}[!t]
\fontsize{9}{11.5}\selectfont
  \caption{
Summary of Finsta-VLM computation cost during the pre-/post-training phase.
}
\vspace{-2mm}
 \setlength{\tabcolsep}{1.5mm}
\begin{center}
  \begin{tabular}{lcrcr}
\toprule

\multicolumn{1}{c}{\multirow{2}{*}{\bf  Item}} & \multicolumn{2}{c}{ \bf  HDVILA} & \multicolumn{2}{c}{\bf Clover}\\
\cmidrule(r){2-3}\cmidrule(r){4-5}
  &\bf Raw  &\multicolumn{1}{c}{\bf +Finsta}  &\bf Raw  &\multicolumn{1}{c}{\bf +Finsta} \\
\hline

\bf Param. (M) & 310 & 397 \textcolor{mred}{\scriptsize{\textbf{($\uparrow$28.1\%)}}} & 258 & 345 \textcolor{mred}{\scriptsize{\textbf{($\uparrow$33.7\%)}}} \\

\bf Data (M)  & 136 & 0.05 \textcolor{nmdeepgreen}{\scriptsize{\textbf{($\downarrow$99.9\%)}}} & 5.3 & 0.05 \textcolor{nmdeepgreen}{\scriptsize{\textbf{($\downarrow$99.1\%)}}} \\

\bf GPU Hours (K) &  174 & 174.35 \textcolor{mred}{\scriptsize{\textbf{($\uparrow$0.2\%)}}} & 65 & 65.24 \textcolor{mred}{\scriptsize{\textbf{($\uparrow$0.37\%)}}} \\

\bf GPU Mem. (G) &  10.6 & 14.9 \textcolor{mred}{\scriptsize{\textbf{($\uparrow$40.6\%)}}} & 9.7 & 13.1 \textcolor{mred}{\scriptsize{\textbf{($\uparrow$35.1\%)}}} \\

\bf GFLOPs & 1,750  & 2,036 \textcolor{mred}{\scriptsize{\textbf{($\uparrow$16.3\%)}}} & 1,424 &  1,653 \textcolor{mred}{\scriptsize{\textbf{($\uparrow$16.0\%)}}} \\

\bottomrule
  \end{tabular}
\end{center}
\vspace{-5mm}
  \label{Efficiency}
\end{table}

\begin{figure*}[t]
 \centering
 \includegraphics[width = 1\textwidth]{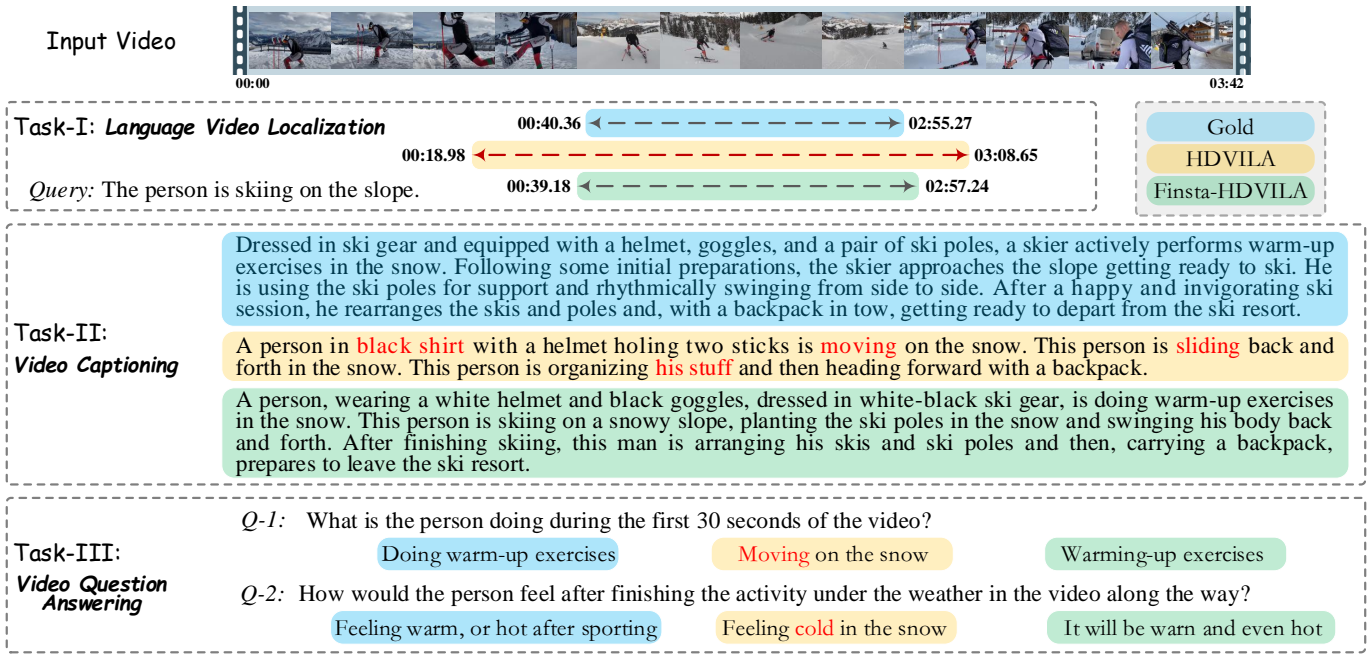}
 \caption{
 Case study, where we present the comparisons on three VL tasks. 
 The red color denotes incorrect predictions.
 }
 \vspace{-3mm}
 \label{fig: case-a}
\end{figure*}

\vspace{-2mm}
\subsection{RQ-4: Is Finsta Computational Efficient?}

Previously we testified that Finsta helps VLMs achieve better end task performance effectively.
Here we try to quantify the computation efficiency of Finsta.
In Table \ref{Efficiency} we summarize HDVILA and Clover VLMs equipped with Finsta, in terms of the computation costs during the pre-/post-training phase.
As seen, since Finsta serves as a module, it introduces additional parameters to the host VLMs, about a total of 87M.
Meanwhile, this takes additional GPU memories.
However, during fine-tuning phase, Finsta has actually been removed from the host VLMs, and thus requires zero additional consumption.
With a play\&play design, it can be conveniently integrated into the existing off-the-shelf VLMs for enhancement without much effort.
That is, one big advantage of Finsta lies in the avoidance of heavy data and GPU time required for post-training.
For example, Finsta-HDVILA reduces 99.9\% data for the training, compared with the original pre-training of HDVILA, where it only costs the increase of 0.2\% training consumption.
Besides, Finsta introduces limited computational burdens, i.e., GFLOPs increase by around 16\%.
This is because the Finsta is designed as a parallel module to the host VLM during inference, which causes no significant delay to the host VLM.
Also Finsta employs Transformer architecture as the backbone, advancing a highly parallel computation of graph data.
Overall, our Finsta system has a cost-effective trade-off between efficacy and efficiency, which is favorable for practical applications.

\vspace{-2mm}
\subsection{RQ-5: How Does Finsta Exactly Advance?}

At last, we try to give a direct understanding of Finsta with respect to how better it succeeds in end tasks.
We empirically present a case study of Finsta's prediction on end tasks.
Representatively, we take the Language Video Localization, Video Captioning and Video Question Answering, where we use a randomly picked testing sample from ActivityNet.
We compare Finsta-HDVILA with HDVILA and the gold annotations in Figure \ref{fig: case-a}.
As observed, while the raw HDVILA makes inaccurate or wrong task predictions, Finsta-HDVILA can give correct predictions that much more coincide with the gold task annotations.
For example, for the Language Video Localization which relies heavily on the video temporal dynamic understanding capability, given the input long video (3 minutes and 42 seconds), HDVILA interprets the query `\emph{The person is skiing on the slope}' by wrongly including many of those video frames that are not relevant to the skiing activity.
In contrast, Finsta-HDVILA determines the action boundary more accurately.
For the captioning and QA, it is key to recognize the video semantics and then generate precise texts, where the strong capabilities of holistic video-language view also with fine-grained cross-model aligning are required.
As seen, the outputs from HDVILA contain inaccurate contents, e.g., failing to detect the event of `\emph{warming-up}', instead recognizing it as `\emph{moving}'.
However, Finsta-HDVILA address all these challenges successfully, yielding high-quality and more correct results.
These cases intrinsically reveal the stronger capabilities of Finsta in enhancing the VL understanding.

\section{Conclusion}\label{sec: conclusion}

In this work, we investigate a fine-grained structural spatio-temporal alignment learning (Finsta) framework for enhancing existing video-language models (VLMs).
First of all, we employ the textual scene graph (TSG) and dynamic scene graph (DSG) to represent the input text and video.
We also unify the TSG and DSG into a holistic scene graph (HSG) with cross-modal coreference edges for bridging two modalities.
Second, we develop a framework based on these SGs, in which a graph Transformer (GTrm) is used to encode the TSG, and also a novel recurrent graph Transformer (R-GTrm) is devised to encode both the DSG and HSG for the spatial-temporal video feature propagation.
We further propose a spatial-temporal Gaussian differential graph Transformer (STGD-GTrm) to strengthen the perception of the changes in objects across spatial and temporal dimensions.
Finally, based on Finsta we perform object-centered spatial contrasting (OSC) alignment and predicate-centered temporal contrasting (PTC) alignment between TSG and DSG respectively, to enhance the VL grounding.
The Finsta system is designed as a plug\&play module, which, via the proposed representation transfer learning, can be conveniently integrated into existing well pre-trained VLMs for further representation augmentation.
We perform extensive experiments on 6 representative VL modeling tasks over 12 datasets in both standard and long-form video scenarios.
Our Finsta framework persistently improves the existing 10 top-performing VLMs and 3 recent LVLMs, and helps push new state-of-the-art VL end tasks significantly in both the fine-tuning and zero-shot settings.
Further in-depth analyses are done to provide a thorough understanding of our system's strengths.

\section*{Acknowledgments}

This work is supported by the National Natural Science Foundation of China (NSFC) Grant (No. 62336008).
This work is also supported by CCF-Baidu Open Fund, NExT Research Center, Skywork AI, Singapore.

\ifCLASSOPTIONcaptionsoff
  \newpage
\fi

\bibliographystyle{IEEEtran}
\bibliography{ref.bib}

\appendices

\section{Illustration of High-order Neighboring}

\begin{figure}[!h]
\begin{center}
\includegraphics[width=0.94\linewidth]{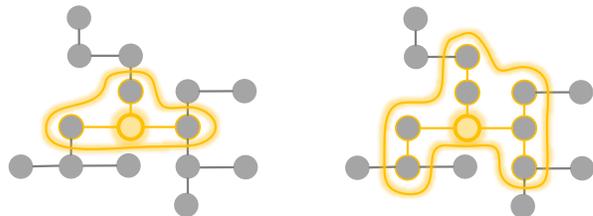}
\end{center}
\caption{
Modeling the high-order neighboring region within the graph.
The glowing yellow nodes are the region center.
}
\label{illus-high-order}
\end{figure}

In our proposed Object-centered Spatial Contrasting (OSC) learning, we consider performing the matching of a high-order region that is centered on any objects.
Because intuitively, a textual object and a visual object should be treated more similarly when the object pair, as well as their modifying contexts (i.e., specific attributes and even relational neighbor objects), are all matched. 
As illustrated in Figure \ref{illus-high-order}, for a TSG object $v^T_i$, we traverse its $n$-th order (e.g., $1$st, $2$nd, or $3$rd) neighbors $\mathcal{N}^T_i$.
And then we obtain the region representation $\overline{h}^T_i$ via pooling operation.
Likewise, for a DSG object $v^D_{t,j}$, we also obtain the $n$-order neighbor representation $\overline{h}^D_{t,j}$.
For the Predicate-centered Temporal Contrasting (PTC) learning, we use the same method to find its $n$-order neighbor spatial region (the representation is marked as $\hat{h}^T_i$) within TSG, as shown in Figure \ref{illus-high-order}.

\section{Details of Video-Language Understanding Tasking}

There has been a series of representative VL modeling tasks, which in our experiments we divide into four groups:
video-to-text transformation (e.g., Video Action Recognition, Video Captioning), text-to-video transformation (e.g., Video-Text Retrieval), VL collaboration (e.g., Video Question Answering) and long-form VL understanding (e.g., Long-Form Video Question Answering, Video-Paragraph Retrieval).
Following we briefly introduce these widely explored tasks, as well as the representative datasets and evaluation metrics.

\subsection{Video Action Recognition}
The task of Video Action Recognition aims to interpret and classify specific actions or activities being performed within the video frames \cite{feichtenhofer2016convolutional}, facilitating the automated semantic understanding of video actions, which is essential for applications in video surveillance, content analysis, and interactive media.
Formally, the task can be modeled as a function \( f: V \rightarrow A \), where \( V \) represents the input space of video sequences and \( A \) denotes the output space of action labels. 
The model \( f \) maps each video \( v \in V \), which is a sequence of frames \( v = [f_1, \cdots, f_n] \), to an action label \( a \in A \). 
The output \( a \) characterizes the predominant action identified in the video sequence.

We consider two representative datasets: Kinetics-400 (K400) \cite{KayCSZHVVGBNSZ17} and Something-Something v2 (SSV2) \cite{GoyalKMMWKHFYMH17}.
K400 contains 400 human action classes, with at least 400 video clips for each action. 
Each clip lasts around 10s and is taken from a different YouTube video. 
The actions are human-focused and cover a broad range of classes including human-object interactions such as playing instruments, as well as human-human interactions such as shaking hands.
The specific data splitting is followed with \cite{Wang0WLZZX0JY22}.
SSV2 is a large collection of labeled video clips that show humans performing pre-defined basic actions with everyday objects. 
It contains 220,847 videos, with 168,913 in the training set, 24,777 in the validation set and 27,157 in the test set. 
There are 174 labels.
Following \cite{Wang0WLZZX0JY22}, we measure the performance in terms of the Top-1 \& Top-5 accuracy.

\subsection{Video Captioning}
Video Captioning involves generating accurate and coherent text that succinctly describes the events or scenes depicted in a video \cite{KrishnaHRFN17}, facilitating easier comprehensibility and accessibility.
Formally, the task can be modeled as \( f: V \rightarrow T \), where \( V \) denotes the input space of video sequences and \( T \) represents the output space of textual captions. 
For a given video \( v \in V \), \( f \) produces a textual caption \( t \in T \), a string of words that encapsulates the key elements and activities presented in the video sequence.

We consider three benchmarks: YouCook2 \cite{ZhouXC18}, MSR-VTT \cite{XuMYR16} and MSVD \cite{chen-dolan-2011-collecting}.
YouCook2 contains 2,000 untrimmed videos from 89 cooking recipes; on average, each distinct recipe has 22 YouTube videos. 
The total video time is 176 hours with an average length of 5.26 mins for each video. Each video captured is within 10 mins and is recorded by camera devices but not slideshows.
The data splitting is followed with \cite{Wang0WLZZX0JY22}.
The standard splits use 6,513 clips for training, 497 clips for validation, and 2,990 clips for testing.
MSR-VTT is a large-scale dataset for open domain video captioning, which consists of 10k video clips from 20 categories, and each video clip is annotated with 20 English sentences by Amazon Mechanical Turks. 
We follow \cite{Wang0WLZZX0JY22}, training models on 9k videos, and reporting results on the 1K-A test set. 
MSVD consists of about 120K sentences collected during the summer of 2010.
Workers on Mechanical Turk were paid to watch a short video snippet and then summarize the action in a single sentence. 
To measure the video captioning performance, we adopt the key metrics of ROUGE-L, CIDEr, METEOR and BLEU@4.

\subsection{Video-Text Retrieval}
The task of Video-Text Retrieval is aimed at retrieving relevant video content based on textual queries \cite{ChenZJW20}.
This task is critical for facilitating the efficient and accurate retrieval of video segments that correspond to specific text-based inputs, especially for applications such as video archiving, digital content management, and online streaming services, etc. 
Formally, this task can be conceptualized as a mapping \( f: T \times V \rightarrow S \), where \( T \) represents the space of textual queries, \( V \) is the set of available videos, and \( S \) denotes the similarity scores or relevance rankings.
For a given textual query \( t \in T \) and a set of videos \( \{v_1, \cdots, v_N\} \in V \), the model \( f \) computes a similarity score \( s_{j} \) for each video \( v_j \), indicating the relevance of \( v_j \) to the query \( t \). The output is a ranked list of videos, ordered by their similarity scores \( S = \{s_1, \cdots, s_m\} \).

LSMDC \cite{MaharajBRCP17}, DiDeMo \cite{HendricksWSSDR17} and also MSR-VTT \cite{XuMYR16} datasets are used in this task.
LSMDC consists of 118K video clips sourced from 202 movies. 
Each video has a caption. 
Evaluation is conducted on a test set of 1K videos. 
When fine-tuning, we sample 2 segments for training and 4 segments for testing and each segment contains 11 frames. 
DiDeMo consists of 10K Flickr videos annotated with 40K sentences. 
We follow \cite{XueHZS00FG22} to evaluate the retrieval performance.
When finetuning, we sample 4 segments for training and 8 segments for testing and each segment contains 11 frames.
MSRVTT contains 10k YouTube sourced videos with 200k text descriptions, where we train the video on 9k videos and evaluate on the rest 1k video.
The evaluation metrics include Recall@1, Recall@5, Recall@10 and MdR.

\subsection{Video Question Answering}
Video Question Answering aims to enable machines to answer questions based on the content of a given video \cite{XuZX0Z0Z17}.
The task requires combining visual comprehension with natural language processing, which is essential for advancing the fields of interactive media, educational technology, etc.
This task can be categorized as Multiple-Choice (MC) and Open-Ended (OE) settings.
For the MC setting, the task can be modeled as \( f: (V, Q, C) \rightarrow A \), where \( V \) represents the set of video sequences, \( Q \) denotes the set of related questions in natural language, \( C \) is a set of candidate answers, and \( A \) is the correct answer from the candidates. 
Given a video \( v \in V \), a question \( q \in Q \) pertaining to the content of the video, and a set of candidate answers \( c \in C \), the function \( f \) aims to identify the correct answer \( a \in C \) that most accurately and relevantly responds to the question \( q \), based on the visual and contextual information present in the video \( v \). 
For OE setting, it can be modeled as a \( f: (V, Q) \rightarrow A \), where \( V \) represents the set of video sequences, \( Q \) denotes the set of related questions in natural language, and \( A \) is the set of possible answers. 
Given a video \( v \in V \), and a question \( q \in Q \) pertaining to the content of the video, \( f \) aims to generate an open-ended answer \( a \in A \), which is typically a word, phrase, or sentence that correctly and relevantly responds to the question \( q \), based on the visual and contextual information present in the video \( v \).

We consider four representative datasets: MSR-VTT-QA \cite{XuZX0Z0Z17}, LSMDC-FiB \cite{MaharajBRCP17}, MSVD-QA \cite{XuZX0Z0Z17} and TGIF-Frame \cite{JangSYKK17}.
MSR-VTT-QA is created based on video and captions in MSR-VTT \cite{XuMYR16}, containing 10k videos and 243k open-ended questions. 
We follow the original work to use an answer vocabulary containing the most common 1.5k answers in the training and validation split as answer candidates. 
For each video, we randomly sample one segment for training and uniformly sample eight segments for testing.
LSMDC-FiB is originally for retrieval task, but reformulated as video QA. 
For the MC setting, it requires the model to find the optimal caption that describes the video out of 5 candidate texts. 
And for the OE setting, it needs to predict a correct word for the blank with a given video and a sentence with blank. 
LSMDC-FiB contains 297k sentences for training and 30k sentences for testing.
Likewise, MSVD-QA is built upon the MSVD data \cite{chen-dolan-2011-collecting}.
We follow the same data pre-processing and spitting as \cite{XuZX0Z0Z17}.
TGIF-Frame is a subset of TGIF \cite{JangSYKK17}, which collects the answerable with just a single frame in the video, and is divided into training set with 35K
questions and test set with 14K questions.
To measure the performance, we adopt the widely adopted accuracy as the metric.

\subsection{Video-Paragraph Retrieval}
Mostly similar to the above Video-Text Retrieval, Video-Paragraph Retrieval aims to retrieve relevant video segments based on a detailed textual description, typically in the long-form texts of paragraphs \cite{zhang2018cross}. 
This is crucial in applications such as digital libraries, content management systems, etc.
The task can be also modeled in the same way as for Video-Text Retrieval.

We use the QuerYD data \cite{OncescuHLZA21} as well as the challenging ActivityNet data \cite{KrishnaHRFN17}.
QuerYD consists of video segments corresponding to textual descriptions, where the full videos are on average 278 seconds long.
Following \cite{OncescuHLZA21}, we use the data with untrimmed videos 1,815 training, 388 developing and 390 testing sets;
and localised descriptions split is 9,113 training, and 1,952 developing and 1,954 testing sets.
ActivityNet includes 20K YouTube untrimmed videos with 100K caption annotations. 
In ActivityNet, each video connects to the descriptions with multiple actions, i.e., over 50\% videos contain more than 4 actions.
The videos are 120 seconds long on average, and most of the videos contain over 3 annotated events with corresponding start/end time and human-written sentences, which contain 13.5 words on average.
This allows to describe multiple events that occur. 
The number of videos in train/validation/test split is 10,024/4,926/5,044, respectively.
Also, the evaluation metrics include Recall@1, Recall@5, Recall@10 and MdR.

\subsection{Long-Form Video Question Answering}
Compared with the aforementioned Video Question Answering, the task of Long-Form one deals with more complex, multi-part queries that require an understanding of prolonged and nuanced videos (e.g., movies), enabling deeper and contextual understanding of long-form video content \cite{SunXS00F22}. 
Formally, the task is modeled in exactly the same manner as for the short one.

We consider the How2QA \cite{li-etal-2020-hero} and VIOLIN \cite{LiuCCGYYL20} datasets.
How2QA contains 44K QA pairs for 22K 60-second clips selected from 9,035 videos.
VIOLIN consists of 95.3K video-hypothesis pairs from 15.9K video clips, spanning over 582 hours of video. 
Each video is paired with 6 statements and is 40 seconds long on average.
The pre-processing and data spitting are followed with \cite{SunXS00F22}.
The accuracy is the main metric.

\section{Configurations of System and Experiments}

\subsection{Specification of Baselines and Backbone VLMs}
For the evaluation of VL tasks, we will compare with different baselines strong-performing or being state-of-the-art on the benchmarks.
We note this type of baselines may not be pre-trained VLMs.
More importantly, we will consider the existing VLMs as our main baselines.
According to the model size, we divide them into the VLMs and LVLMs.
Also different VLMs may either have a dual-stream-sum architecture with three `\emph{text-video-multimodal}' encoders, or with certain encoder(s) being absent.
Different (L)VLMs are pre-trained on different amounts of corpus, and with different volumes of parameters.
Following we briefly describe them.
Also in Table \ref{VLMs-sum} we give a snapshot of the (L)VLMs in terms of their characteristics.

\begin{table}[!t]
\fontsize{9}{12}\selectfont
\setlength{\tabcolsep}{2mm}
\begin{center}
\caption{
A summary of (L)VLMs used as our backbones.
For the architecture, T: textual encoder, V: video encoder, C: cross-modal encoder.
Params: the amount of VLM parameters.
Data: the amount of data for VL pre-training.
For LVLMs (with $*$), strictly speaking, the textual encoder also serves as the cross-modal encoder.
}
\label{VLMs-sum}
  \begin{tabular}{lccccc}
\toprule

\multicolumn{1}{c}{\multirow{2}{*}{\bf Model}}&   \multicolumn{3}{c}{\bf Architecture} & \multicolumn{1}{c}{\multirow{2}{*}{\bf Params}} & \multicolumn{1}{c}{\multirow{2}{*}{\bf Data}} \\
\cmidrule(r){2-4}
& {T} &  {V} & {C} & & \\

\midrule
\multicolumn{5}{l}{\bf $\bullet$ VLM}\\

\quad UniVL \cite{luo2020univl} & \checkmark  & \checkmark  & \checkmark  & 183M & 1.2M\\

\quad VIOLET \cite{abs-2111-12681} & \checkmark  & \checkmark  & \checkmark  & 198M & 11.8M\\

\quad VideoCLIP \cite{xu-etal-2021-videoclip}  & \checkmark  & \checkmark  & \ding{55}  & 203M &  1.1M \\

\quad CLIP4Clip \cite{LuoJZCLDL22} & \checkmark  & \checkmark  & \ding{55}  & 86M & 380K \\

\quad MCQ \cite{GeGLLSQL22}  & \checkmark  & \checkmark  & \checkmark  & 168M &  5.8M \\

\quad HDVILA \cite{XueHZS00FG22} & \checkmark  & \checkmark  & \checkmark  & 310M & 136M\\

\quad Clover \cite{abs-2207-07885} & \checkmark  & \checkmark  & \checkmark  & 258M & 5.3M \\

\quad LFVILA \cite{SunXS00F22} & \checkmark  & \checkmark  & \checkmark  & 277M & 8.5M\\

\quad All-in-one \cite{wang2023all} & \ding{55}  & \ding{55}  & \checkmark  & 110M & 3.72M\\

\quad InternVideo \cite{wang2022internvideo} & \checkmark  & \checkmark  & \checkmark  & 1.3B & $\sim$100M\\

\hline
\multicolumn{5}{l}{\bf $\bullet$ LVLM}\\

\quad Video-LLaMA \cite{abs-2306-02858}$^*$ & \halfcheckmark  & \checkmark  & \checkmark  & 7.6B & 2B \\

\quad Video-ChatGPT \cite{maaz2023video}$^*$ & \halfcheckmark  & \checkmark  & \checkmark  & 7.4B & 32B \\

\quad Video-LLaVA \cite{lin2023video}$^*$ & \halfcheckmark  & \checkmark  & \checkmark  & 7.4B & 32B\\

\bottomrule
  \end{tabular}
\end{center}
\end{table}

\vspace{1mm}
\noindent\textbf{$\blacktriangleright$ VLMs}, (mostly) with million-level parameters:
\begin{compactitem}
    \item \textbf{UniVL}\footnote{\url{https://github.com/microsoft/UniVL}} is a unified VL pre-training model which has a standard text\&video encoder linking to the corss-model encoder \cite{luo2020univl}.

    \item \textbf{VIOLET}\footnote{\url{https://github.com/tsujuifu/pytorch_violet}} takes a fully end-to-end VL Transformer architecture, with a Masked Visual-token Modeling (MVM) pre-training task for video modeling \cite{abs-2111-12681}.

    \item \textbf{VideoCLIP}\footnote{\url{https://github.com/facebookresearch/fairseq/tree/main/examples/MMPT}}.
    Xu et al. \cite{xu-etal-2021-videoclip} present a contrastive approach to pre-train a VLM model that comes with only the text and video encoders.
    
    \item \textbf{CLIP4Clip}\footnote{\url{https://github.com/ArrowLuo/CLIP4Clip}}. Luo et al. \cite{LuoJZCLDL22} exploit the pre-trained CLIP for video clip retrieval.

    \item \textbf{MCQ}\footnote{\url{https://github.com/TencentARC/MCQ}}.
    Get et al. \cite{GeGLLSQL22} enable VLM with fine-grained VL interactions via a Multiple Choice Question pretext task, where a cross-modal module BridgeFormer is proposed.

    \item \textbf{HDVILA}\footnote{\url{https://github.com/microsoft/XPretrain/tree/main/hd-vila}} is 
    built based on the text\&video and multimodal Transformers that enforce interactions of the learned video features with diversified texts.

    \item \textbf{Clover}\footnote{\url{https://github.com/LeeYN-43/Clover}}.
    Huang et al. \cite{abs-2207-07885} propose a correlated video-language pre-training model, which improves cross-modal feature alignment and fusion via a novel tri-modal alignment pre-training.

    \item \textbf{LFVILA}\footnote{\url{https://github.com/microsoft/XPretrain/tree/main/LF-VILA}}.
    Sun et al. \cite{SunXS00F22} propose a long-form video-language pre-training model, which is trained on a large-scale long-form video and paragraph dataset with a multimodal temporal contrastive learning loss to learn the temporal relation across different modalities.

    \item \textbf{All-in-one}\footnote{\url{https://github.com/showlab/all-in-one}}.
    Wang et al. \cite{wang2023all} introduce an end-to-end VLM that embeds raw video and text into joint space using a unified cross-modal backbone architecture. Since without the separate text\&video encoders, the VLM has comparatively fewer parameters.

    \item \textbf{InternVideo}\footnote{\url{https://github.com/OpenGVLab/InternVideo}} also takes the dual-stream-sum architecture with three `\emph{text-video-multimodal}' encoders, while further explores masked video modeling and VL contrastive learning as the pre-training objectives \cite{wang2022internvideo}.

\end{compactitem}

\vspace{1mm}
\noindent\textbf{$\blacktriangleright$ LVLMs}, where a video encoder connects to a language-centered LLM to understand the input video signals, with billion-level parameters:
\begin{compactitem}
    \item  \textbf{Video-LLaMA}\footnote{\url{https://github.com/DAMO-NLP-SG/Video-LLaMA}}. Zhang et al. \cite{abs-2306-02858} employ a video Q-former as the video encoder, and use Vicuna-7B \cite{vicuna} as the LLM for understanding both the video and language.

    \item \textbf{Video-ChatGPT}\footnote{\url{https://github.com/mbzuai-oryx/Video-ChatGPT}} \cite{maaz2023video} leverages the CLIP-L/14 visual encoder to extract both spatial and temporal video features, and then fed into a linear layer, which projects the video features into the Vicuna-7B (v1.1) LLM.

    \item \textbf{Video-LLaVA}\footnote{\url{https://github.com/PKU-YuanGroup/Video-LLaVA}} \cite{lin2023video} takes LanguageBind \cite{zhu2023languagebind} to align the representations of videos to a unified vision-language feature space, which are then mapped to the LLM, Vicuna-7B (v1.5).
    
\end{compactitem}

\begin{table}[!t]
\fontsize{9}{12}\selectfont
 \setlength{\tabcolsep}{1.1mm}
\begin{center}
  \caption{
A summary of post-training datasets.
}
  \label{post-data-summary}
  \begin{tabular}{lccc}
\toprule

\multicolumn{1}{c}{\multirow{2}{*}{\bf Aspect}}&   \multicolumn{2}{c}{\bf Normal (short) Videos} & \multicolumn{1}{c}{\bf Long-form Videos} \\
\cmidrule(r){2-3}\cmidrule(r){4-4}
& {\small WebVid} &  {\small HD-VILA} & {\small LF-VILA}\\
\midrule
 Raw amount (K)  & 2,500 & 103,000 & 8,000 \\
 Used amount (K)  & 25 (1\%) & 25 (0.025\%) & 30 (0.375\%)\\
\hline
 Sen. num. (M)  & 2.5 & 103 & 8.5 \\
 Sen. len. (token.)  & 12.0 & 32.5  & 307.9 \\
\hline
 Vid. num. (M) & 2.5 & 103 & 8.5 \\
 Vid. len. (Sec.)  & 18.0 & 13.4 & 100.2 \\

\bottomrule
  \end{tabular}
\end{center}
\end{table}

\subsection{Specification of Post-Training}.

The post-training of our system takes a warm-up process.
We first train the Finsta alone on the text-video pairs with TSG and DSG annotations for 2 epochs as warming-up, using a batch size of 300. 
We use AdamW \cite{LoshchilovH19} optimizer with a weight decay 5e-3 and betas (0.9, 0.98).
The learning rate is first warmed-up by 1 epochs to 5e-3 and then decays.
When the Finsta tends to converge, we then perform the knowledge distillation, and inject the Finsta representations into the host VLM as described above.
All trainings are conducted on 16 NVIDIA A100 GPUs.

For the post-training of Finsta-VLMs on the normal (short) form videos, we use a total of 50K VL pairs, with 25K sampled from WebVid-2.5M \cite{BainNVZ21} and 25K sampled from HD-VILA-100M \cite{XueHZS00FG22}.
The WebVid dataset is a substantial and diverse collection of internet-sourced video clips, each paired with textual annotations for a comprehensive video-language understanding. 
It spans a broad spectrum of categories, ensuring a rich sample of visual content that is matched with high-quality descriptions. Primarily used for advancing video-language models, the dataset supports a range of tasks such as captioning, search, and question answering.
The HD-VILA dataset is an expansive video-language dataset curated for high-definition video understanding and associated language tasks. It consists of 100 million video-text pairs, with high-resolution video clips paired with detailed textual descriptions. 
The dataset is designed to enable fine-grained analysis and foster advancements in video understanding, captioning, and retrieval, by providing a diverse array of content spanning multiple domains. 
For the post-training of Finsta-LFVILA\footnote{We denote the Finsta-LFVILA post-trained on the normal (short) form videos as \emph{S-Vid}, and on the long-form data as \emph{L-Vid}.}, we further consider the use of long-form data as what LFVILA has been trained, i.e., with 30K VL pairs sampled from LF-VILA-8M \cite{SunXS00F22}, where the avg. video duration is 100.2 sec. and the avg. text length is 307.9 tokens.
In Table \ref{post-data-summary} we summarize the characteristics of the data for Finsta-VLM post-training.
In this regard, we can note that the Finsta-VLM post-training consumes quite significantly fewer VL pairs, compared with the pre-training for the host VLM.
For example, HDVILA is pre-trained with 136M VL pairs.

\subsection{Scene Graph Parsing}

For the TSG annotations, we mainly follow the prior practice of SG applications.
We also perform filtering to remove objects, relations, and attributes that appear less than 5 times in all the parsed scene graphs. 
After such filtering, we obtain 7,021 objects, 2,256 relations, and 4,895 attributes in TSGs. 
To obtain the DSGs for videos, we first perform keyframe extraction.
For each video, we flexibly extract 10-50 keyframes while preserving their order.
The key process of DSG parsing has been elaborated in Section 3 in the main article.
We filter the object, attribute, and relation annotations by keeping those that appear more than 500 times in the training set, which helps screen the less informative noises.
The resulting vocabulary of visual SG contains 776 objects, 187 attributes, and 95 relations in DSGs.

For the experiment of studying the influence of SG parsing quality, we designed the experiment as follows.
For the DSG part, the DSG generation task consists of localizing a set of objects, classifying their category labels as well as predicting relations between each pair of these objects.
Since we cannot change the FasterRCNN on object detection, we consider varying the performance of the scene graph generation (SGG) step in SG parsing.
For the textual SG generation, we vary the performance of the first step of dependency tree parsing (a.k.a., SGG).
We vary these parser qualities by training them with early existing, such that they will show different developing performances.

\section{Extended Experimental Analyses}

In this section, we show several analyses further that are not presented in the main article.


\begin{figure}[t]
\centering
\includegraphics[width=0.98\columnwidth]{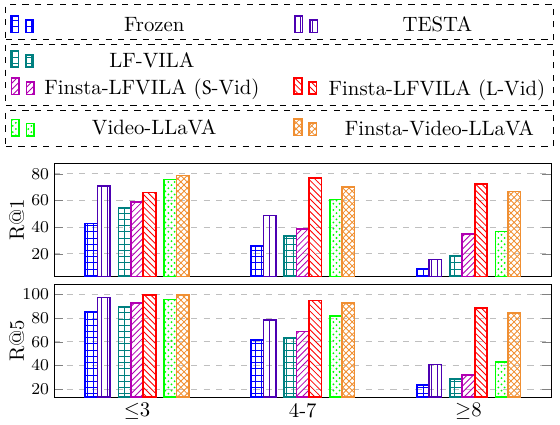}
\caption{
Video-Paragraph Retrieval results with varying numbers of video actions in ActivityNet data.
}
\label{fig:Complex-Actions}
\end{figure}

\subsection{Performance in Handling Multiple Video Actions}

We test whether our system enhances video-language understanding by sufficiently modeling the temporal dynamics.
To reach this, we directly evaluate the capability to understand the videos with complex and lengthy dynamics.
We consider the most challenging video-paragraph retrieval data, ActivityNet \cite{KrishnaHRFN17}, where each paragraph contains multiple numbers of actions described in the corresponding long video, which can be seen as the complex dynamics scenario.
Specifically, we divide the data into three groups, with one containing less than or equal to 3 actions, and one from 4 to 7 actions, and one no less than 8.
We measure the retrieval performance in terms of R@1 and R@5.
By observing the results in Figure \ref{fig:Complex-Actions}, we can see that when the action number increases, the overall performance of baselines drops dramatically.
However, with Finsta, backbone VLMs gain very striking improvements, especially for the very complex videos (i.e., $\ge$8).
For example, Finsta-LFVILA (L-Vid) improves LFVILA by 48.3\% R@1 and 58.3\% R@5 for videos with $\ge$8 actions.
This evidently suggests that, our Finsta especially enhances the video and language understanding with complex temporal dynamic scenes.

\begin{figure}[!t]
\begin{center}
\includegraphics[width=0.92\columnwidth]{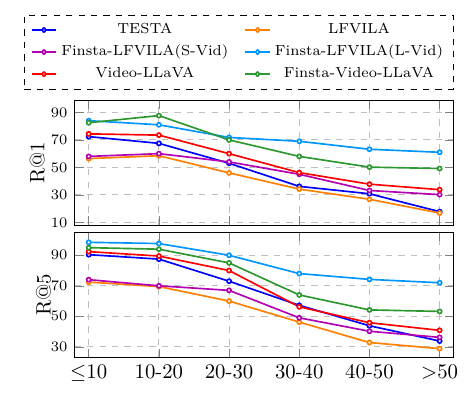}
\end{center}
\caption{
Video-Paragraph Retrieval performance under different input text lengths (words) in ActivityNet data.
}
\label{app-length}
\end{figure}

\begin{figure}[!t]
\begin{center}
\includegraphics[width=1\columnwidth]{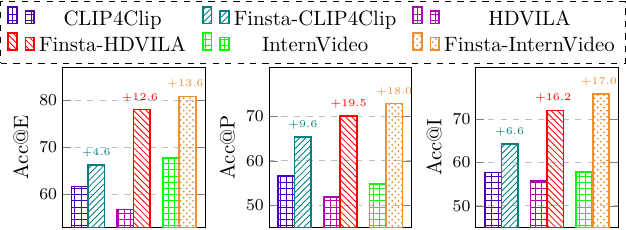}
\end{center}
\caption{
Performance on Causal-VidQA in terms of E: \emph{video explanation}, P: \emph{video prediction}, I: \emph{video imagination}. 
}
\label{collaboration}
\end{figure}

\subsection{Performance in Handling Longer Sentences}

Essentially, lengthy sentences and videos require a stronger capability of temporal dynamics modeling.
As DSG structural representation well depicts the video dynamics, Finsta is expected to better advance this aspect.
Here we take a further step, and evaluate how better Finsta handles longer content.
We perform Video-Paragraph Retrieval experiments with different VLMs under varying lengths of input query texts.
We again use the ActivityNet data, where the average length of paragraphs is 50.4 words, and the shortest text is over 20 words. 
We additionally add some video-text pairs with sentences shorter than 20 words from the DiDeMo data.
In Figure \ref{app-length} we show the results.
As seen, with the increase in text length with complex temporal dynamics, the retrieval performance drops quickly.
Yet, Finsta helps much effectively counteract the growth of content length, than the vanilla LFVILA and Video-LLaVA (L)VLMs.
This is especially clear when handling longer text queries, i.e., with length $>$30.

\subsection{Strengthening Video-Language Collaboration}

Directly measuring the VL collaboration can be tricky; alternatively, we take an indirect manner, described as follows.
Video and language actually come with distinct while complementary semantics towards one thing, i.e., intuitively texts bring abstract expressions, and videos provide concrete \& intricate features.
For example, from language, `\emph{a man is sad}'; from vision, `\emph{a man is sobbing}'.
When VLMs learn a good VL collaboration, they can comprehend the same object/event from either a visual signal or textual signal that is implicitly linked to each other.
Thus enabling VLMs to perceive these two input modalities collaboratively is an indispensable prerequisite for a deep VL understanding.
Under such assumption, we verify our system with an advanced video QA benchmark, Causal-VidQA \cite{li2022representation}, where the videos and text questions often do not have explicit coreference, and advanced high-level VL understanding abilities are required, such as 
1) explaining the intentions of actions and the procedures to certain targets;
2) predicting the future action beyond current video clips;
3) imagining what would happen under different conditions.
In Figure \ref{collaboration} we plot the comparing results.
As seen, all the VLMs boosted by Finsta witness significant enhancement in advanced VL understanding, indicating stronger VL collaboration.
Notedly, the CLIP4Clip that has an architecture without corss-model encoder receives the least improvement, since the VL collaboration modeling in Finsta's HSG encoder is not able to be injected to CLIP4Clip.

\begin{table*}[t]
\centering
\caption{Performance of Finsta-HDVILA trained using incomplete/impaired SG annotations.}
\fontsize{9}{11}\selectfont
\setlength{\tabcolsep}{3.5mm}
\begin{tabular}{llcccc}
\hline
\multicolumn{2}{c}{\bf SG Sources} & \bf \specialcell{K400\\{\scriptsize(VAR, Top-1)}} & \bf \specialcell{MSR-VTT\\{\scriptsize(VC, M)}} & \bf \specialcell{DiDeMo\\{\scriptsize{(VTR. R@1)}}}	 & \bf \specialcell{MSVD\\{\scriptsize(VQA, Acc.)}}  \\   
\midrule

 \multicolumn{2}{l}{HDVILA (w/o Finsta\&SG)} & 	78.6 & 	32.4 & 	28.8	 & 50.7 \\
 \hline
 
\multicolumn{2}{l}{With high-quality SG} & 	83.4 & 	36.9 & 	49.3 & 	53.3 \\
 \hdashline

\multirow{3}{*}{With impaired SG} & Node missing rate: 10\% & 81.9	 & 35.2 & 	44.5 & 	51.5 \\
 & Node missing rate: 30\% & 80.7 & 	33.1 & 	35.0 & 	50.9 \\
 & Node missing rate: 50\% & 73.3 & 	30.5 & 	25.1 & 	46.3 \\
 
\hline
\end{tabular}
\label{tab:SG-annotation-incomplete}
\end{table*}

\begin{table*}[t]
\centering
\caption{Statistics of TSG \& DSG generated via different approaches.}
\fontsize{9}{11}\selectfont
\setlength{\tabcolsep}{4.5mm}
\begin{tabular}{lcccccc}
\hline
& \multicolumn{3}{c}{ \bf  TSG} & \multicolumn{2}{c}{ \bf  DSG} \\
\cmidrule(r){2-4}\cmidrule(r){5-7}
 &\bf  object & \bf 	relation & \bf 	attribute & 	\bf object & \bf 	relation & \bf 	attribute \\

\midrule
Specialist SG Parser & 	7,021 & 	2,256 & 	4,895 & 	776	 & 187 & 	95  \\   
LLM Parser & 	13,364 & 	6,571 & 	7,630 & 	6,357 & 	1,911 & 	4,350  \\   

\hline
\end{tabular}
\label{tab:SG-annotation-llm-labels}
\end{table*}

\begin{table*}[t]
\centering
\caption{Performance of Finsta-HDVILA trained using different SG annotations.}
\fontsize{9}{11}\selectfont
\setlength{\tabcolsep}{5mm}
\begin{tabular}{lcccc}
\hline
\multicolumn{1}{c}{\bf SG Sources} & \bf \specialcell{K400\\{\scriptsize(VAR, Top-1)}} & \bf \specialcell{MSR-VTT\\{\scriptsize(VC, M)}} & \bf \specialcell{DiDeMo\\{\scriptsize{(VTR. R@1)}}}	 & \bf \specialcell{MSVD\\{\scriptsize(VQA, Acc.)}}  \\   
\midrule

 \multicolumn{1}{l}{HDVILA (w/o Finsta\&SG)} & 	78.6 & 	32.4 & 	28.8	 & 50.7 \\
 \hline
 
\multicolumn{1}{l}{Specialist SG Parser} & 	83.4 & 	36.9 & 	49.3 & 	53.3 \\
 \hdashline

LLM SG Parser & 83.0 & 	36.2 & 	50.1 & 	54.5 \\

\hline
\end{tabular}
\label{tab:SG-annotation-llm}
\end{table*}

\subsection{Extended Discussion on Impact of SG Annotation}

\subsubsection{Influence of Incomplete SG Annotations}

In real scenarios, SG representations often contain noise, incomplete even impaired annotations during the generation processing.
Thus, It is important to study the reliance of our system on such scenarios.
Here, we design an experiment setting to simulate such case.
We randomly discard some nodes from the TSG, DSG, and HSG, with the dropping rate at 10\%, 30\% and 50\%, resulting in incomplete/impaired SGs. 
We then post-train Finsta with this data of incomplete SG annotations to observe the changes in Finsta's performance on end tasks. 

The results are shown in the Table \ref{tab:SG-annotation-incomplete}.
It is clear that the quality of SG annotations significantly impacts the post-training phase. 
A decrease in SG quality inevitably affects Finsta's ability to learn fine-grained cross-modal representations from SGs. 
This finding and the conclusion are mutually consistent with the study in the main article for studying the influence of SG parsing quality.
However, we would like to emphasize that, in most cases, the SG parsers we currently rely on can provide reliable SG annotations. 
Most importantly, fortunately, our Finsta system relies on SG annotations only during the post-training phase, meaning the quality of SG parsing does not impact the performance during inference on downstream tasks. This significantly reduces the influence of SG annotation quality on Finsta's efficacy.

\subsubsection{Influence of SG Parsing with LLMs}

It is a question, i.e., whether it is feasible to use an existing multimodal LLM for SG parsing, rather than an SG parser that is trained on the in-domain training set. 
It is possible that SG annotations derived from a multimodal LLM could potentially be more robust, given the LLM's stronger open vocabulary and better generalization capabilities, which could lead to more accurate or factually consistent SG annotations. 
This, in turn, might help the Finsta system to better assist video LLMs. 
To explore this, we conduct an evaluation experiment during this period. 
Our approach involves training our system using the same training methodology on an identical number of SG-annotated V-L pairs. 
However, in contrast to the method where we used an SG parser, the SG annotations for this experiment were derived from a variety of open-source generators. 
Specifically, we employed SoTA text LLMs (Flan-T5 XXL) for producing TSGs and SoTA video LLMs (video-LLaVA) for generating DSGs. 
All LLMs are prompted to generate DSG structures through few-shot instructions. 
The quantity of SGs generated is the same, totaling 50K TSG and DSG pairs for V-L analysis.

We first compare the node type labels of SG annotations produced by the LLM-parser and the specialist SG parser, as shown in Table \ref{tab:SG-annotation-llm-labels}.
It's evident that the SGs generated by LLMs possess richer label information, undoubtedly benefiting from the open vocabulary generalization capabilities of LLMs. This indeed confirms the reviewer's speculation.
Following this, we conduct post-training on our system using this version of SG-annotated data and compared its performance on end tasks. 
We primarily focus on the performance of the Finsta-HDVILA backbone across four tasks.
It can be observed from Table \ref{tab:SG-annotation-llm} that the performance of the Finsta system trained on SG data generated by LLMs is very close to, and in some cases even slightly higher than, that of the Finsta system trained on SG data parsed by specialists. 
Yet, both significantly outperform the performance of the vanilla HDVILA backbone itself. 
This indicates that although SGs parsed by LLMs may contain noise or randomness, as long as they incorporate a reasonable structural fine-grained representation, they can still provide substantial support for downstream VLM tasks when integrated with our Finsta system. 
Furthermore, due to the additional label diversification stemming from LLMs, it may even help achieve better results.

\begin{table*}[t]
\fontsize{9}{11.5}\selectfont
\setlength{\tabcolsep}{1.mm}
\begin{center}
\caption{
List of the top static and dynamic objects automatically learned via the STGD-GTrm modules.
For dynamic objects, we also show the most common associated predicates (actions) behind each dynamic object.
}
\begin{tabularx}{\linewidth}{lX}
\hline
 \multicolumn{1}{c}{\bf Type} & \multicolumn{1}{c}{\bf Object Labels} \\
\hline
\textbf{Static objects} & house, building, tree, road, bench, mountain, river, sky, book, phone, computer, desk, chair, lamp, bed, pillow, blanket, closet, playground, cup, plate, fork, knife, spoon, bottle, glass, stove, refrigerator, microwave, painting, window, door, wall, floor, ceiling, roof, fence, sidewalk, street, field, clock, watch, calendar, map, sign, poster, blackboard, whiteboard, bookshelf, fireplace, shower, bathtub, toilet, sink, towel, soap, shampoo, conditioner, toothbrush, garden, snow, flower, bush, grass, rock, sand, dirt, clay, mud, gravel, statue, fountain, bridge, tunnel, path, trail, highway, alley, crosswalk, parking lot, curtain, sofa, carpet, shelf, picture, plant pot, notebook, pencil, camera, television, remote, heater, air conditioner, broom, dustpan, light switch, outlet, refrigerator magnet, oven, toaster, blender, coffee maker, dish washer \\ 
\hline
\textbf{Dynamic objects} & car$|$\emph{drive}, person$|$\emph{walk}, athlete$|$\emph{run}, airplane$|$\emph{fly}, vehicle$|$\emph{run}, bicycle$|$\emph{ride}, motorcycle$|$\emph{speed}, ball$|$\emph{play}, boat$|$\emph{sail}, ship$|$\emph{cruise}, truck$|$\emph{deliver}, runner$|$\emph{jog}, bird$|$\emph{fly}, cat$|$\emph{prowl}, kid$|$\emph{scream}, dog$|$\emph{bark}, horse$|$\emph{gallop}, elephant$|$\emph{trample}, lion$|$\emph{roar}, fish$|$\emph{swim}, whale$|$\emph{breach}, skateboard$|$\emph{skate}, dolphin$|$\emph{leap}, kangaroo$|$\emph{hop}, rabbit$|$\emph{bounce}, snake$|$\emph{slither}, lizard$|$\emph{climb}, frog$|$\emph{jump}, butterfly$|$\emph{flutter}, bee$|$\emph{buzz}, spider$|$\emph{spin}, skateboarder$|$\emph{grind}, skier$|$\emph{slalom}, swimmer$|$\emph{dive}, dancer$|$\emph{twirl}, child$|$\emph{play}, thief$|$\emph{steal}, driver$|$\emph{park}, writer$|$\emph{type}, climber$|$\emph{ascend}, shopper$|$\emph{browse}, protester$|$\emph{chant}, traveler$|$\emph{explore}, deer$|$\emph{graze}, crow$|$\emph{caw}, monkey$|$\emph{swing}, squirrel$|$\emph{scamper}, owl$|$\emph{hoot}, butterfly$|$\emph{land}, ant$|$\emph{carry}, mosquito$|$\emph{buzz}, bear$|$\emph{forage}, fox$|$\emph{hunt}, eagle$|$\emph{soar}, octopus$|$\emph{swim}, turtle$|$\emph{swim}
 \\

\hline
\end{tabularx}
\label{tab:obj}
\end{center}
\end{table*}

\section{Extended Case Study}

We try to give a direct understanding of Finsta with respect to how it exactly works and helps VLMs achieve better VL learning.
To this end, we consider empirically conducting some case studies.
At the end of the main article, we empirically present a case study of Finsta’s prediction on end tasks. 
Here, we show two more empirical case analyses. 
First, we show the top-frequent static objects and dynamic objects automatically induced from the STGD-GTrm module.
And then, we visualize the fine-grained spatial and temporal grounding between TSG and DSG in Finsta.

\subsection{Automatic Static\&Dynamic Object Detection}

We design an STGD-GTrm encoder to enable the DSG and HSD encoder, R-GTrm, to better perceive the changes in objects across spatial and temporal dimensions.
Meanwhile, STGD-GTrm perceives the difference between nodes in moving or being stationary. 
Since Finsta can make more accurate predictions, intuitively the system is able to automatically recognize those objects of \emph{static object} and \emph{dynamic object} from DSG during the representation learning, with which we also find their counterpart nodes from TSG.
By analyzing the distribution densities of different objects in $\kappa(v_i^{\Delta t})$ (cf. Eq 30), we can interpret the different types of objects.
We summarize all these objects, and record their frequencies.
In Table \ref{tab:obj} we representatively show the top-frequent static objects and dynamic objects automatically detected, where the dynamic objects are also associated with the frequent predicates.
It is surprising to find that the recognized objects of both static and dynamic types tend to be quite accurate, coinciding with the factual conditions.
For example, the objects from the static group, such as `{house}', `{building}', `{tree}', `{road}', `{park}', `{bench}', all tend to play the background roles that may not actively move in the video.
And for the objects as well as their paired actions from the dynamic group, such as `car$|$\emph{drive}', `person$|$\emph{walk}', `athlete$|$\emph{run}', `airplane$|$\emph{fly}', `vehicle$|$\emph{run}', `bicycle$|$\emph{ride}', they mostly tend to serve the foreground roles in videos that observably move.

\begin{figure*}[t]
 \centering
 \includegraphics[width = 1\textwidth]{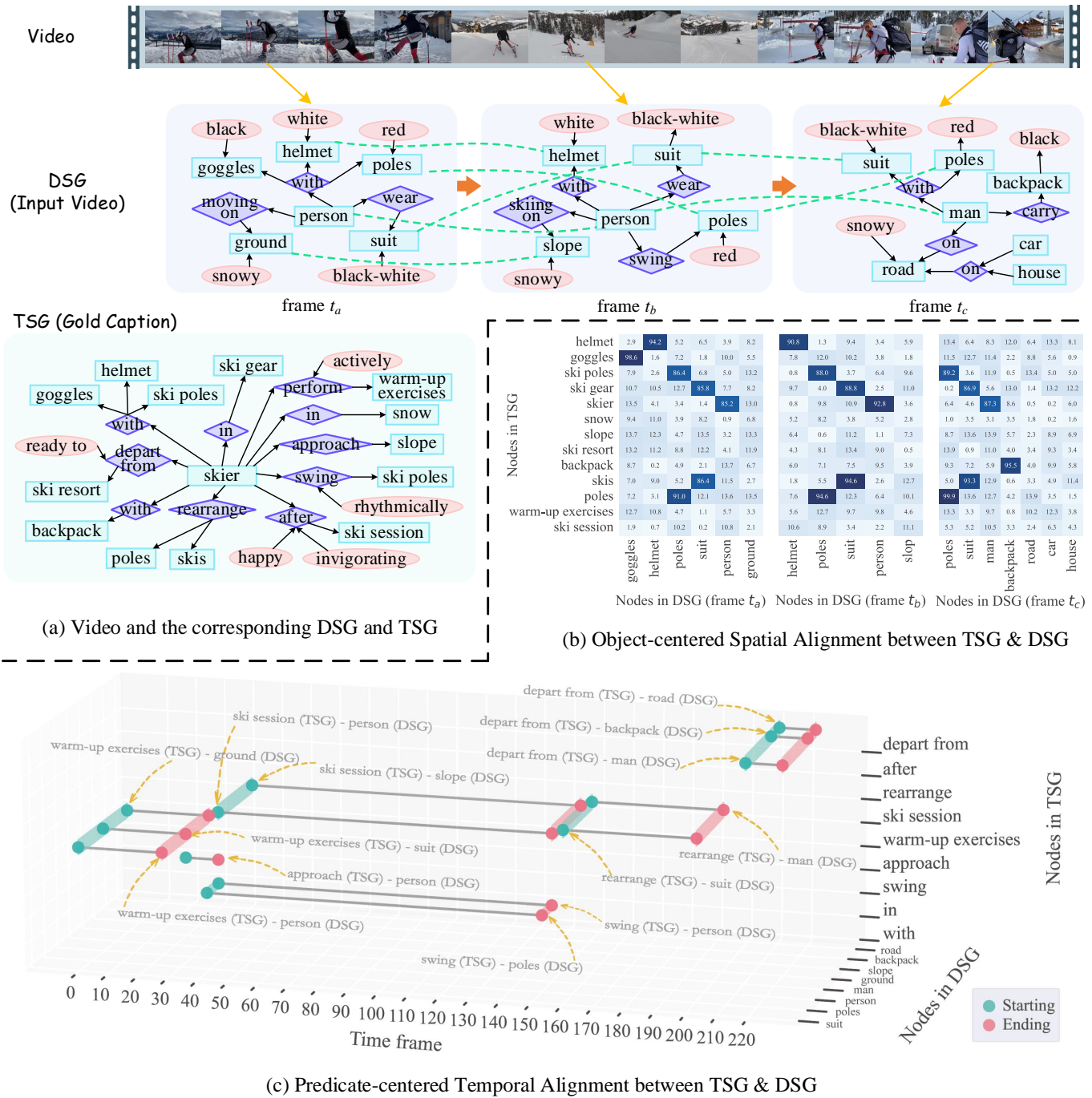}
 \caption{Visualization of the fine-grained object-centered spatial alignment and predicate-centered temporal alignment between TSG and DSG.
 Due to the space limitation, we only visualize three (non-consecutive) frames of DSG.
 }
 \vspace{-3mm}
 \label{fig: case-b}
\end{figure*}

\subsection{Visualization of Fine-grained Spatial and Temporal Grounding in Finsta}

Finally, we try to visualize the intrinsic working mechanism of how Finsta achieves enhanced fine-grained spatial and temporal grounding/alignment.
Using the same instance as in Figure 16 in the main article, we exhibit the DSG of the input video and the TSG of the gold caption encoded within the Finsta.
As shown in Figure \ref{fig: case-b} (a), both the constructed DSG and TSG are informative enough to depict the raw video and text information.
For the DSG, the temporal coreference edges are also accurate.
Further, we try to visualize the grounding between TSG and DSG in terms of the object-centered spatial alignment (OSC) and the predicate-centered temporal alignment (PTC), respectively.
We interpret the bipartite similarity in $S^o_{i,t,j}$ (cf. Eq 37) and $S^p_{i,t:t+m,j}$ (cf. Eq 39) for the two alignments, and visualize them in Figure \ref{fig: case-b} (b) and (c), respectively,
We can see that for the spatial-level static object alignment, textual nodes and visual nodes have a very precise correspondence.
Further we check the temporal-level dynamic object alignment, where the TSG and DSG nodes are temporally grounded in the time axis with time duration explicitly shown.
Interestingly, the system quite accurately finds the correspondence of actions between language and video in the temporal dimension.
For example, for the `\emph{swing}' action (predicate) in TSG, Finsta accurately finds its temporal counterparts in video DSG that depict the same action, including `\emph{person}' and `\emph{poles}' objects, which has been precisely grounded on the time order and duration.

\end{document}